\documentclass{article}

\usepackage{arxiv}

\usepackage[utf8]{inputenc} % allow utf-8 input
\usepackage[T1]{fontenc}    % use 8-bit T1 fonts
\usepackage{hyperref}       % hyperlinks
\usepackage{url}            % simple URL typesetting
\usepackage{booktabs}       % professional-quality tables
\usepackage{amsfonts}       % blackboard math symbols
\usepackage{nicefrac}       % compact symbols for 1/2, etc.
\usepackage{subcaption}
\usepackage{graphicx}
\usepackage{multicol}
\usepackage{multirow}

\usepackage{microtype}      % microtypography
\usepackage{lipsum}
\renewcommand{\vec}[1]{\mathbf{#1}}
\newcommand{\code}[1]{\texttt{#1}}

\title{ClusterFlow: how a hierarchical clustering layer make allows deep-NNs more resilient to hacking, more human-like and easily implements relational reasoning}

\author{
  Ella M. Gale \\
  School of Chemistry\\
  University of Bristol\\
  Bristol\\
  \texttt{ella.gale@bristol.ac.uk} \\
   \And
 Oliver J. Matthews\\
  \\
  \\
  \\
  \texttt{ojm@codersoffortune.net} \\
}

\begin{document}
\maketitle

\begin{abstract}
Despite the huge recent breakthroughs in neural networks (NNs) for artificial intelligence (specifically deep convolutional networks) such NNs do not achieve human-level performance: they can be hacked by images that would fool no human and lack `common sense'. It has been argued that a basis of human-level intelligence is mankind's ability to perform relational reasoning: the comparison of different objects, measuring similarity, grasping of relations between objects and the converse, figuring out the odd one out in a set of objects. Mankind can even do this with objects they have never seen before. Here we show how ClusterFlow, a semi-supervised hierarchical clustering framework can operate on trained NNs utilising the rich multi-dimensional class and feature data found at the pre-SoftMax layer to build a hyperspacial map of classes/features and this adds more human-like functionality to modern deep convolutional neural networks. We demonstrate this with 3 tasks. 1. the statistical learning based `mistakes' made by infants when attending to images of cats and dogs. 2. improving both the resilience to hacking images and the accurate measure of certainty in deep-NNs. 3. Relational reasoning over sets of images, including those not known to the NN nor seen before. We also demonstrate that ClusterFlow can work on non-NN data and deal with missing data by testing it on a Chemistry dataset. This work suggests that modern deep NNs can be made more human-like without re-training of the NNs. As it is known that some methods used in deep and convolutional NNs are not biologically plausible or perhaps even the best approach: the ClusterFlow framework can sit on top of any NN and will be a useful tool to add as NNs are improved in this regard. 
\end{abstract}

% keywords can be removed

\section{Introduction}

%\paragraph{The aim} of this work is to demonstrate that the addition of the ClusterFlow framework to pre-trained deep neural networks (NNs) makes these NNs behave in a more humanlike manner (in this case the human is a 4month infant). This is computer science, the goal is to improve machine learning algorithms not to say anything new about psychology results (the psychology is there as a benchmark for the algorithm).

Relational reasoning has been suggested as the basis of all intelligence. Here we present a novel algorithm that is capable of adding relational reasoning to convolutional neural networks (CNNs), increases their resilience to hacking and demonstrates similar outcomes to a human being on surprise reactions familiar and novel stimuli.

\subsection{Spaces}

\begin{figure}
    \centering
    \includegraphics[width=\textwidth]{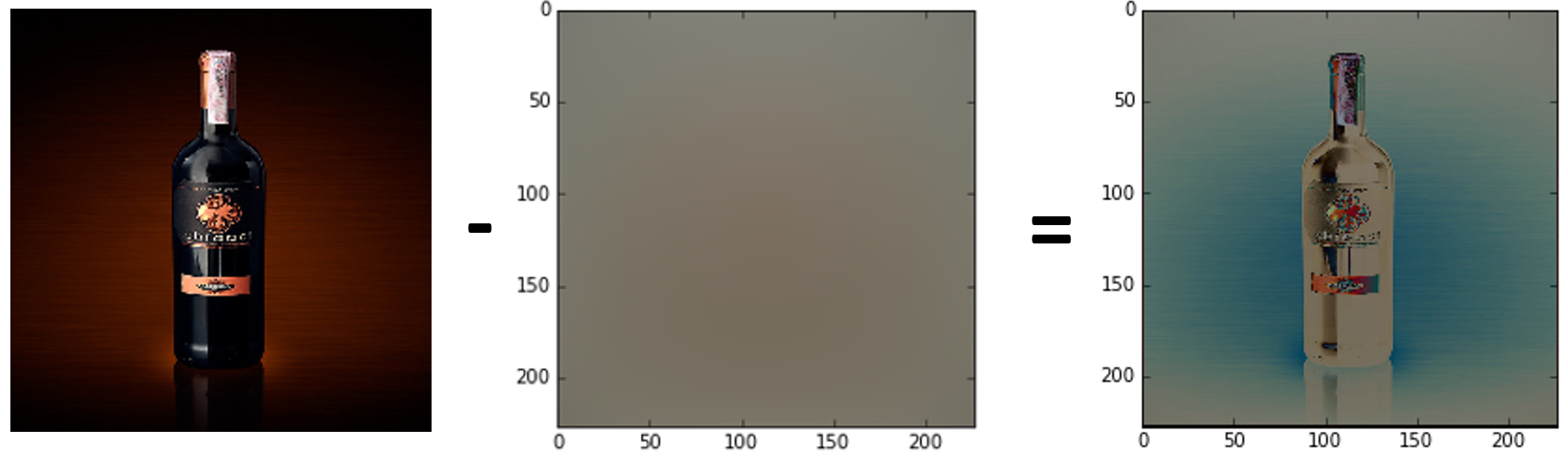}\\
    \begin{tabular}{p{5cm}p{6cm}p{5cm}}
        a. Wine bottle in pixel space $x(\Omega_x)$, & b. ImageNet average, & c. normalised wine bottle $x'(\Omega_x)$ \\
    \end{tabular}
    \includegraphics[width=0.47\textwidth]{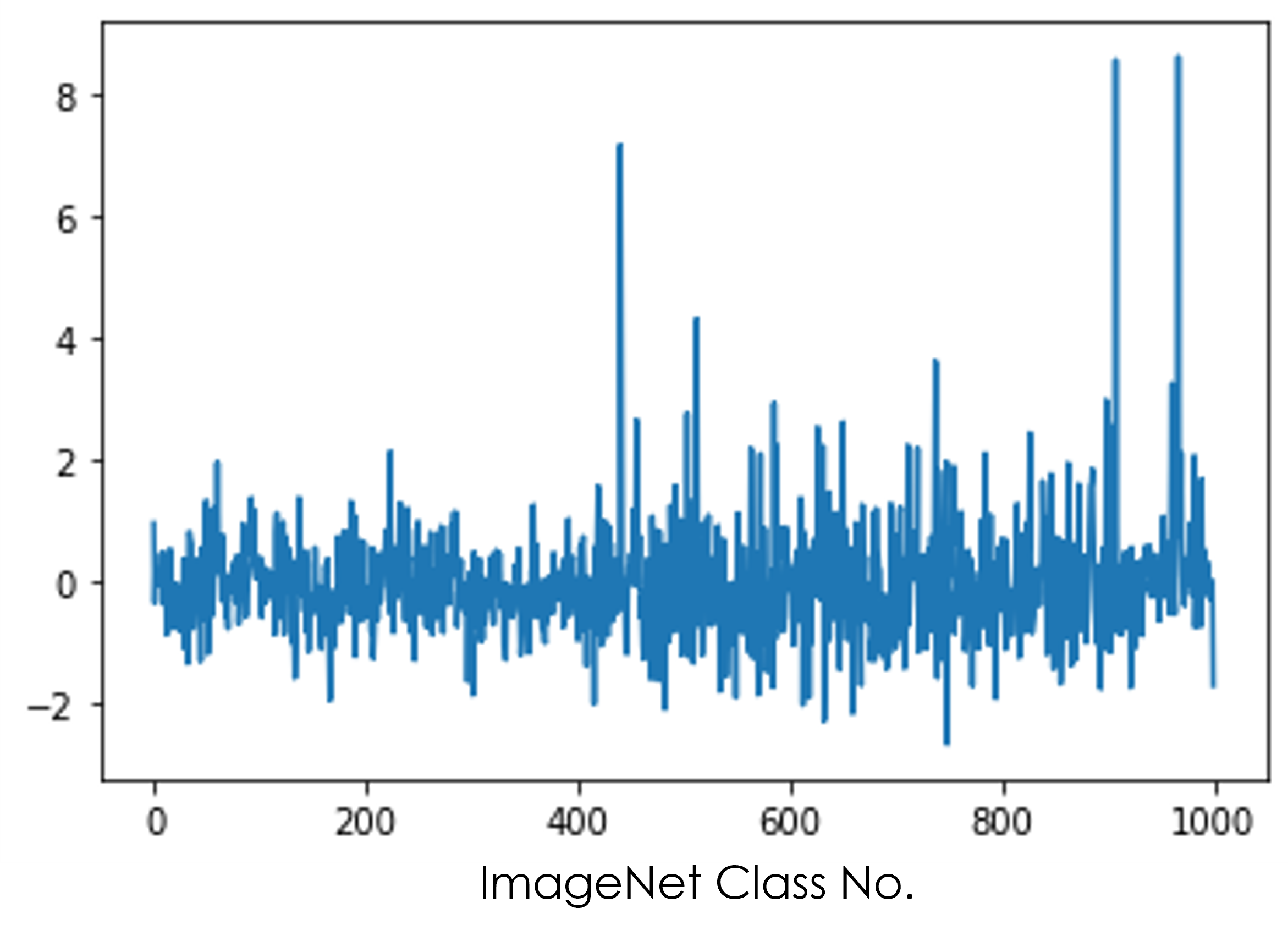}
    \includegraphics[width=0.51\textwidth]{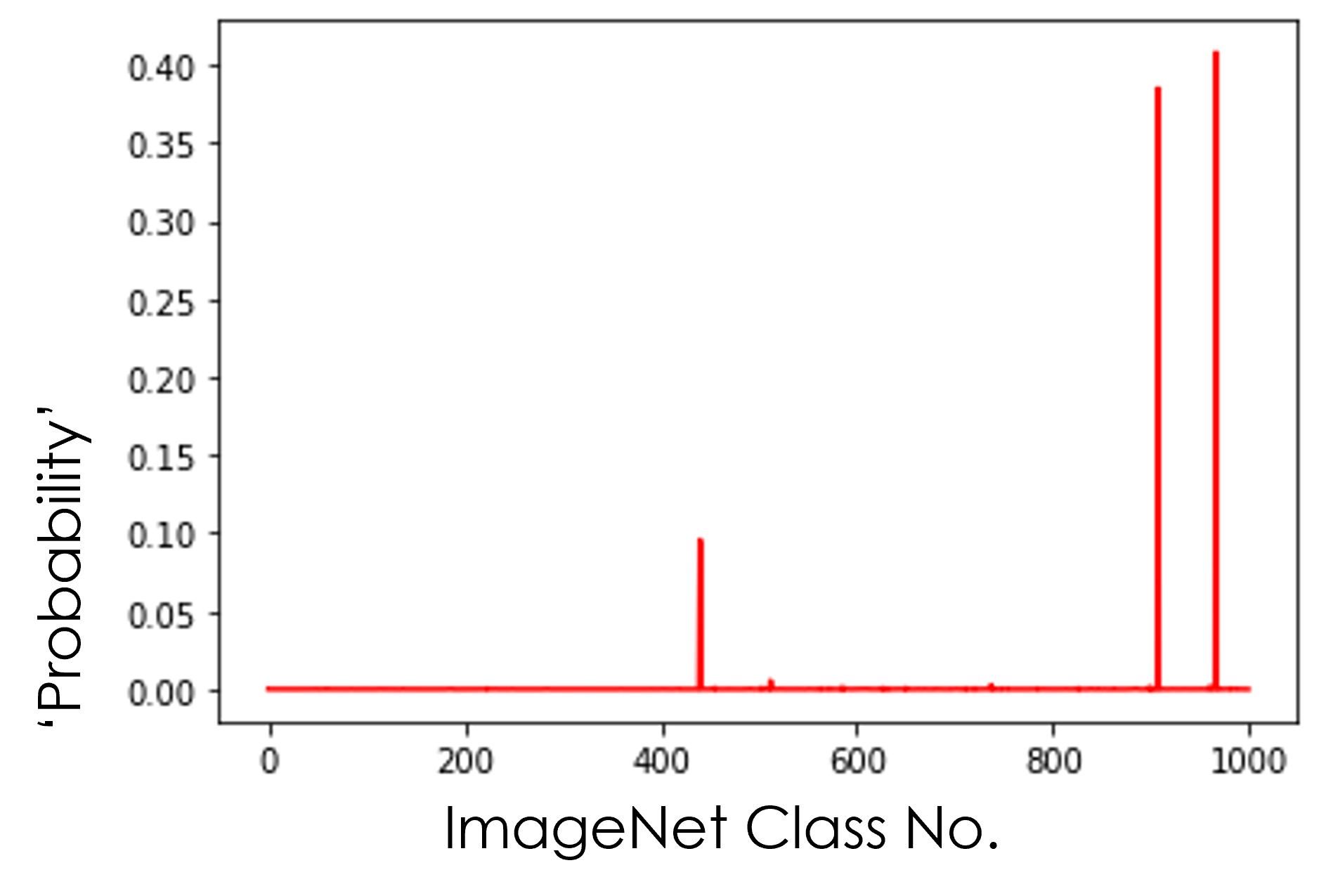}\\
    \begin{tabular}{p{8cm}p{8cm}}
    c. Wine bottle in class space, $\vec{y}(\Omega_c)$ &d. Wine bottle in `probability' space $\vec{y}(\Omega_\rho)$\\
    \end{tabular}
    \includegraphics[width=0.8\textwidth]{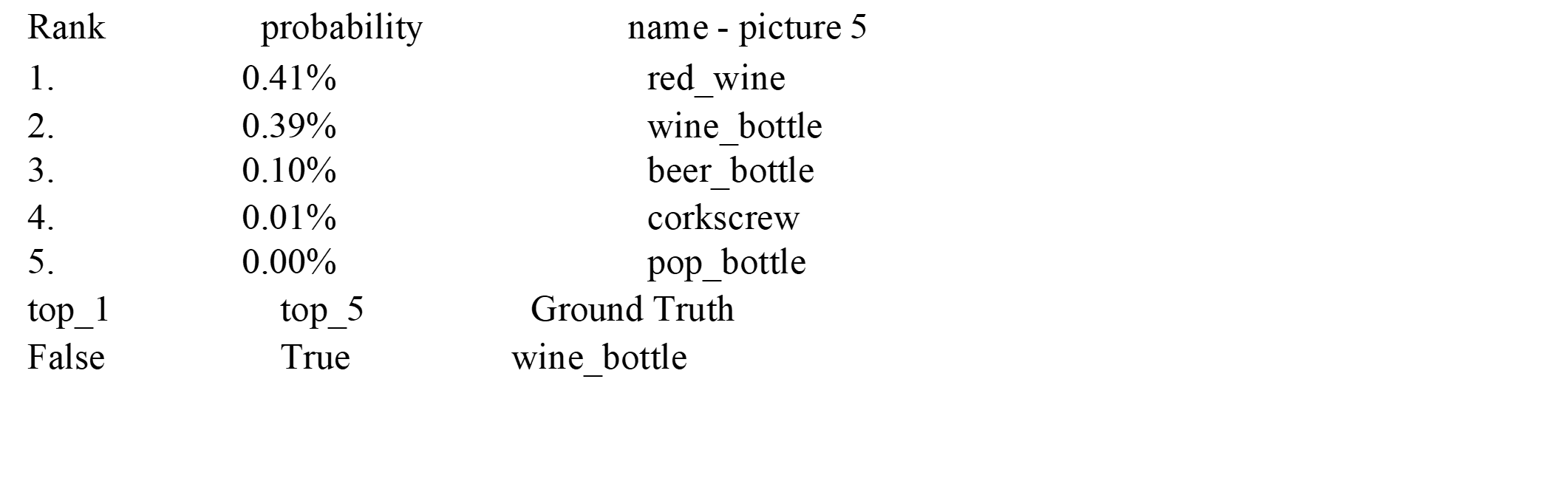}\\
    e. Output, containing the probability $y(\Omega_\rho)$, the human readable names $y(\Omega_h)$ and the ground truth $\hat{y}(\Omega_h)$.
    \caption{Different embeddings of the input and output data used in a NN}
    \label{fig:wine_zB}
\end{figure}

Machine learning can generally be thought of as mapping from an input space, $X$ (signals/data), to an output space $y$ (labels). It is known that neural networks (NNs) work as function approximators, and we assume that there exists a function, $y=f(x)$, that can do this mapping. For example, if we wanted a NN that could identify alcohols from structures, the function could be: `is R-OH present, if yes, y=True, if no, y=False'. The domain, $\Omega$ is the space that $X$ is in, for example, $X(\Omega_x)$, is the image data $X$ in the domain of a pixel grid, $\Omega_x$, several examples of $x(\Omega_x)$ are given in figure~\ref{fig:wine_zB}a and c, where $x$ is a picture of wine bottle. The neural network learns a hypothesis function
\begin{equation}
    F(X(\Omega_x)) = y(\Omega_y),
\end{equation}
which relates signals to labels.\cite{bronstein2021geometric}\footnote{Note that people rarely write the domain in in these equations, but there is always a domain.} In this work, the labels and domains they are embedded in is crucial, see table~\ref{tab:y} for a summary. For classification tasks we generally have a SoftMax layer, $S$, which implements equation~\ref{eq:SoftMax}. Specifically, the output after SoftMax is a commonly referred to as probability (although we argue that it is not a fair probability, see section~\ref{ssec:SoftMax}), and thus the output $y$ in the domain of `probability', $\Omega_\rho$, as given by equation~\ref{eq:top}, 

\begin{equation}
    y(\Omega_\rho) = S(y(\Omega_c))
    \label{eq:top}
\end{equation}
and this the output of the top of the final neural network (alternatively, called the output or the output of the `prob' layer). An example of $y(\Omega_\rho) |_{\mathrm{wine bottle}}$ is given in figure~\ref{fig:wine_zB}d. Below the SoftMax layer is the `class' layer which is as wide as the number of classes, $N_{Class}$, and gives a vector of $N_{Class}$, the range of which depends on the activation function of the layer below that, for example $y(\Omega_c)$ can be $(-\infty, +\infty)$ for relus or $[0,1]$ for sigmoidal activations. The equation for calculating the class vectors is given in equation~\ref{eq:class_vect}

\begin{equation}
    y(\Omega_c) = F(X(\Omega_p))\;.
    \label{eq:class_vect}
\end{equation}

These $y(\Omega_c)$ are also called latent variables, or bottleneck activations and these would be the middle layer of an autoencoder.  An example of $y(\Omega_c)|_{wine bottle}$ is given in figure~\ref{fig:wine_zB}c. It is common in neural network research to take a pre-trained CNN and remove the SoftMax layer and then add a new output layer(s) and retrain the top layers of the CNN, and this is called a decapitated network and is described by equation~\ref{eq:class_vect}. This is commonly done to leverage a large dataset for training on the general problem, before specialising with a smaller dataset. The first training can be unsupervised for example, or a large, deep pre-trained image recognition network can be repurposed for a specific image recognition problem. In this work, we develop and use the ClusterFlow algorithm to cluster over the class vectors $\vec{y}(\Omega_c)$. We take these class vectors as a vector space $\vec{Y} = \{ \vec{y}\}$ embedded in the learned latent space manifold, $\Omega_c$. This `Euclidization of data' is a very important, as we later use the distances. The CF algorithm technically converts this data to a metric space, as we add a metric to the system as well. We also have the human readable labels embedded in $\Omega_h$, the ImageNet label space $\Omega_l$ and the binary codes that result from a 1-hot encoding of these labels which are embedded in $\Omega_b$ -- it is these that directly relate to the probability space as the CNN is trained it tries to minimise $y(\Omega_\rho) - \hat{y}(\Omega_b)$, see table~\ref{tab:y} for a summary.

\begin{table}[htp]
    \centering
    \begin{tabular}{|c|c|p{6cm}|c|}
    \hline
    Outputs & Type of space & Relevant information & Example \\
    \hline
        $y$ & $y=F(X)$ & Guessed label & \\
    \hline
        $y(\Omega_{\rho})$ & `Probability' space & Sums to 1 & [0.001, 0.9, 0.099]\\
        $y(\Omega_{c})$ & Class space & Related to activation function & [-10, 44.6, 20]\\
        \hline
        $\hat{y}$ & & Ground truth / true label & \\
        \hline
        $\hat{y}(\Omega_{h})$ & Human-readable label space & Natural language & `Daisy'\\
        $\hat{y}(\Omega_{l})$ & ImageNet label space & $\hat{y}(\Omega_{h}) = Dict[\hat{y}(\Omega_{l})]$& `n01914609'\\
        $\hat{y}(\Omega_{b})$ & 1-hot encoding & Sums to 1 & [0,1,0]\\
        \hline
    \end{tabular}
    \caption{Labels and label spaces}
    \label{tab:y}
\end{table}

\subsection{Hacking CNNs}

\begin{figure}[htp]
    \centering
    \includegraphics[width=0.5\textwidth]{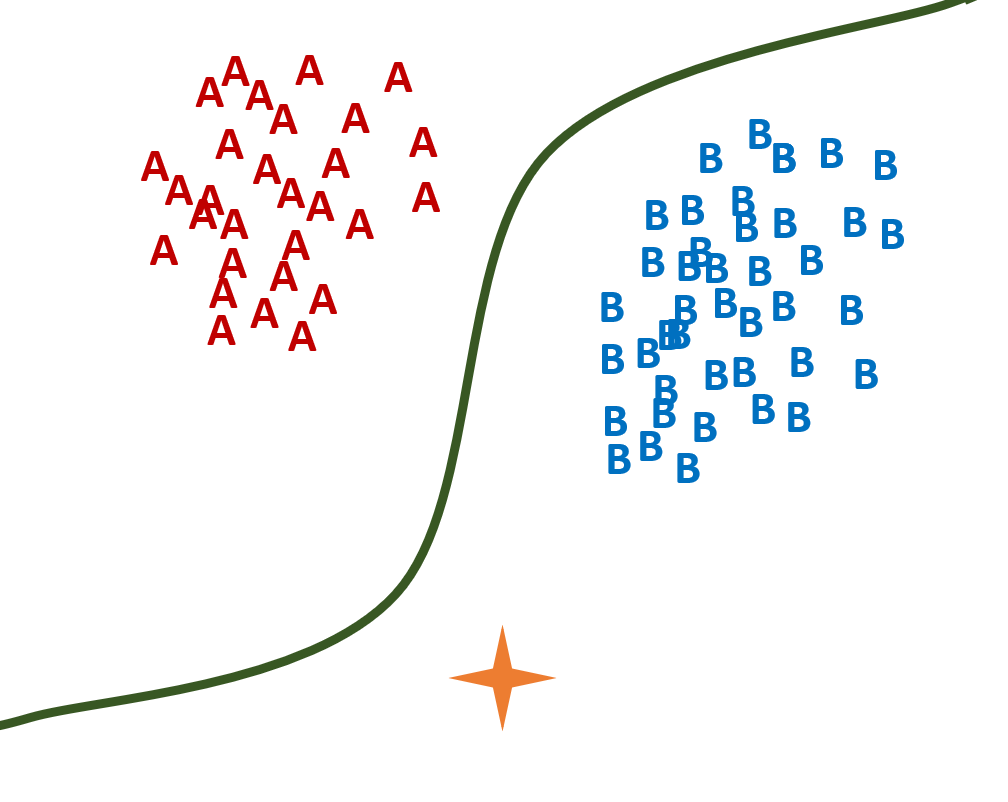}\\
    a. A NN divides class space ($y(\Omega_c)$) into regions using a hypersurface (green). Hacking attacks (orange) that is far from anything seen before, is on the correct side of the to be assigned to class B with high confidence.\\
    \includegraphics[width=0.5\textwidth]{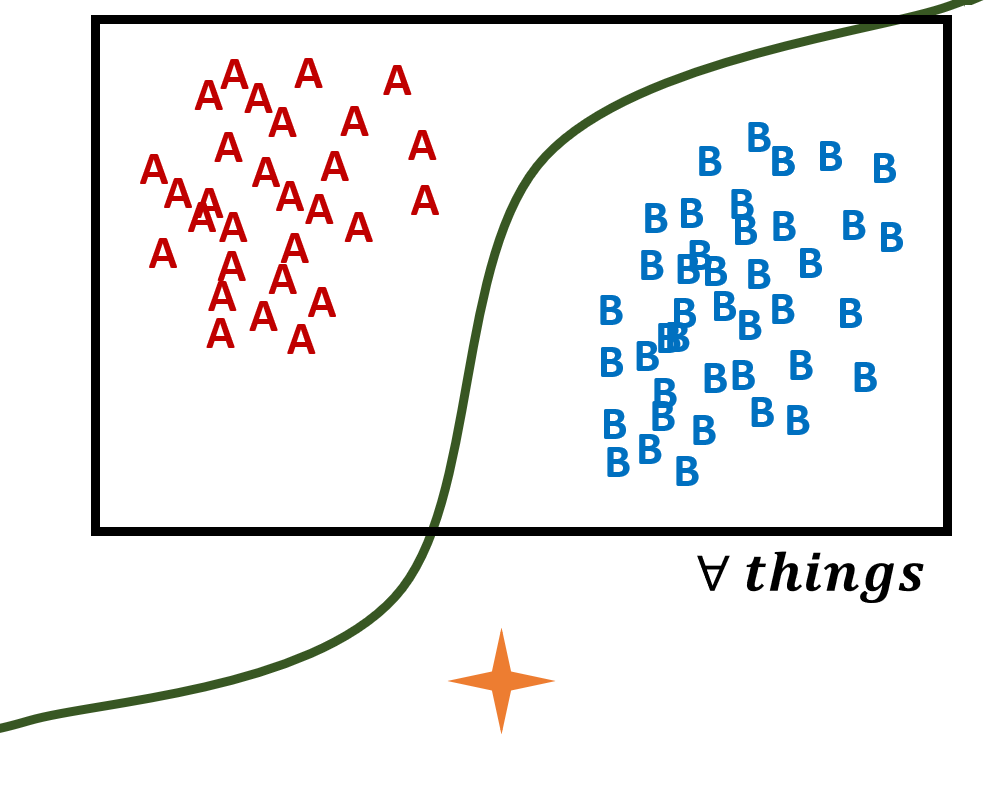}\\
    b. ClusterFlow instead encapsulates world knowledge in a space of all things seen so far (black box).\\
    \includegraphics[width=0.5\textwidth]{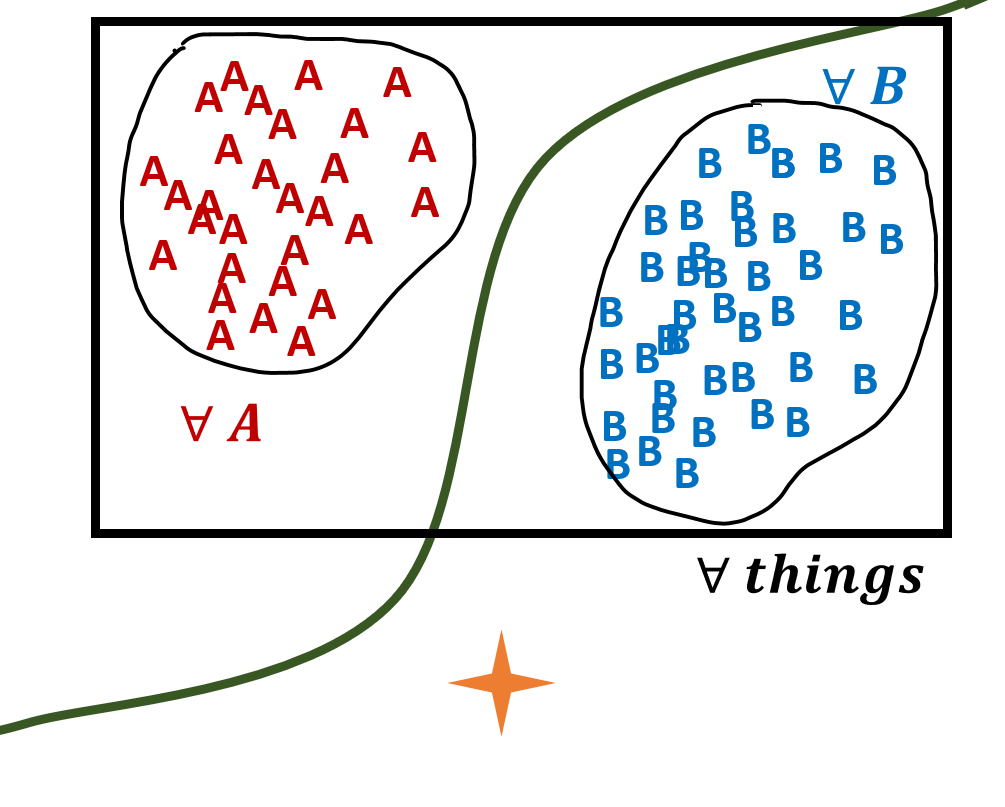}\\
    c. ClusterFlow can further subdivide this space into regions that contain all examples of the classes seen.
    \caption{How ClusterFlow works to avoid hacking attacks.}
    \label{fig:general_idea}
\end{figure}

In 2015 it was found that CNNs could be easily `hacked', where nonsense images were classified with very high confidence (more than 99\%), for examples see figure~\ref{fig:fooling_zB}.\cite{40}These images were created by evolutionary algorithms to try to `hack' CNNs. It was discovered that adding a small amount of colour `noise'\footnote{It's not actually noise, but looks like it to the human eye}, which makes an image indistinguishable from the original for a human could also alter the classification of an image.\cite{goodfellow2014explaining} Or a small, highly detailed sticker could mislead the CNN into ignoring the larger class.\cite{61}  Some of this may be caused by the fact that CNNs attend to small, high-frequency features rather than large, low-frequency features, which is how humans identify pictures.\cite{41-baker2018deep} And this is due to convolutions and weight-sharing to increase translational invariance, the CNNs lose gross shape features. Another cause could be the way that NNs work. A NN takes in high-dimensional information and converts it to a lower dimensional (but still high dimensional) representation in its latent variables. A fascinating aspect of NNs is that they usually also convert this data to a `nice' representation, nice as the surface is generally a lower dimensional manifold in the latent variable space, it is smooth, possibly differentiable, and this is the kind of structure favoured by physical laws, so our understanding and tools work well with it. To classify A from B a NN learns a manifold (hypersurface) that divides latent variable space ($y(\Omega_c)$) into regions for each class, see figure~\ref{fig:general_idea}a. Thus examples like the orange star that are far outside anything that has been seen before like the orange star in figure~\ref{fig:general_idea} are then categorized by being in a class (in this case, B) with high probability. ClusterFlow was invented to try to solve this problem as an entry to the Google-sponsored Kaggle competition on both hacking and defending against hacking of CNNs. The main idea is to encapsulate \textit{world knowledge} in the space of all things seen so far by literally marking out a subspace in the latent variable space. A point like the orange point in figure~\ref{fig:general_idea}b is then marked as being outside of world knowledge and then given a correspondingly low confidence (NaN in the case of ClusterFlow). The interiour of the subspace of all things seen can then be clustered into regions for each class, as in the subspace for all A and for all B in figure~\ref{fig:general_idea}c, and so ClusterFlow can be used to more conservatively give the correct labels with a reduced confidence for unusual or hacking attacks. 

In this paper, we will use cluster flow as a NN layer on top of pre-trained CNNs to demonstrate that it can increase the resilience of these CNNs to hacking.

% there are more references

\subsubsection{SoftMax and biased probability\label{ssec:SoftMax}}

The SoftMax equation is given by:

\begin{equation}
    y = \frac{e^x}{\sum_x e^x} = S(x)
    \label{eq:SoftMax}
\end{equation}

When using SoftMax, researchers commonly cite this paper\cite{53} and state that the output of the SoftMax layer is a probability for the following reasons: A probability, $p$, must be between 0 and 1: $0 < p \leq 1$; $p$ must be positive valued $p>0$.
%\begin{enumerate}
%    \item A probability, $p$, must be between 0 and 1: $0 < p \leq 1$
%    \item $p$ must be positive valued $p>0$
%\end{enumerate}
I posit that to be a (true) probability, then the measurement must be fair. In paper\cite{53}, all the output units were sigmoidal, so $\vec{y}(\Omega_c) = [y_1, y_2, y_3, ... y_n]$ where $n$ is the number of classes, all the $y_x$ could only take values in the range [0,1]: i.e. $0 \leq y_x \leq 1$. So, we could take the $y_x(\Omega_c)$ as a measure of probability that class $x$ is present (with the assumption that a higher activation is associated with the NN having more evidence that that class is present, there is evidence for this assumption\cite{GaleObjectDetectors}), and then by comparing these probabilities with each other, you can state that the class with the highest $y(\Omega_c)$ is the one that the NN is most confident is in the picture. Then, converting these numbers to `prob' space ($y(\Omega_{\rho})$) by using SoftMax is just a convenience to normalise the output. However, by moving away from sigmoidal activation units to relu and elu we break this link between probability and the $y(\Omega_c)$ level. For example, a relu has the range $[0, \infty)$. We posit that the NN uses the freedom in the output range to separate similar classes, and as the NN is trained to identify the training set, it is not trained to give a confidence measure. The confidence measure is put in afterwards as an assumption. But if the classes do not use roughly the same range in output to encode their likeliness, then this assumption treats the classes differently, i.e, there is a bias introduced whereby classes with higher activation ranges (seen during training) are highly favoured in confidence. As this is not treating the classes equally, the SoftMax output is not a fair probability. If we make the assumption that the range of the activations of the training set at the end of training is the range of confidence the NN has. Imagine we have an image which is class a, but the NN is uncertain, and gives it $y_a(\Omega_c) =5$, $y_b(\Omega_c) =5.5$ and $y_c(\Omega_c) =6$, for three similar classes. As shown in table~\ref{tab:SoftMax_zB}, the SoftMax operation gives class $c$ twice as much confidence than class $a$, so we can see how confidence could be inflated for classes which have higher ranges. Also, if $0\leq y_a(\Omega_c) \leq 8$ and $0\leq y_c(\Omega_c) \leq 30$, we can now see that perhaps the NN is more certain about class a, as if we divide these numbers by the range we get  0.625 for a and 0.2 for c (and these normalised values are now a probability). So, SoftMax can inflate confidence and perhaps benefit classes with a high range at the cost of those with a low range. Other workers have argued against the use of SoftMax\cite{57} and suggested similar solutions to the problem of over-confidence.\cite{56} Due to this largely hidden bias, SoftMax is an unfair probability.

\begin{table}
    \centering
    \begin{tabular}{|c|c|c|c|c|}
    \hline
        Domain & & $a$ & $b$ & $c$ \\
    \hline
        $x$ & $\Omega_c$ & 5.0 & 5.5 & 6.0 \\
        $\frac{x}{\sum x}$ & & 0.32 & 0.33 & 0.34 \\
        $\frac{e^x}{\sum_x e^x}$ &$\Omega_\rho$ & 0.18 & 0.31 & 0.51 \\
    \hline
    \end{tabular}
    \caption{SoftMax converts from class space to so-called `probability' space. As it uses exponents, classes with very similar scores are pulled apart. $\vec{x} = [a, b, c]$.}
    \label{tab:SoftMax_zB}

\end{table}

\begin{figure}[htp]
    \centering
    \includegraphics[width=\textwidth]{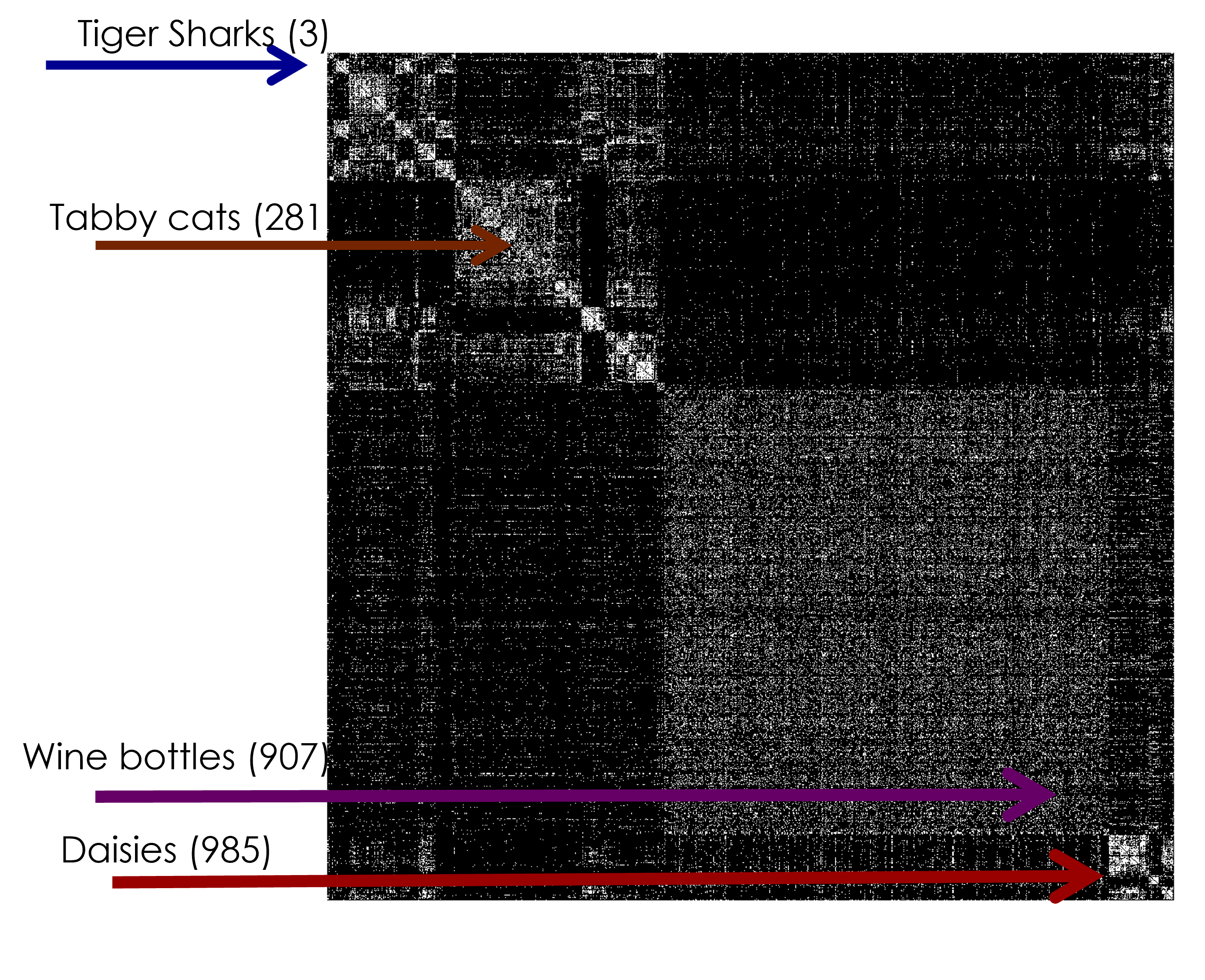}
    \caption{2-confuser matrix for AlexNet on ImageNet data. The four classes chosen for ImageNet\_4 are in different confuser clusters.}
    \label{fig:confuser}
\end{figure}

However, a good point of SoftMax is that class space is under-trained and contains very rich descriptions of the input, see for example the class space vector in figure~\ref{fig:wine_zB}c. These class space vectors represent the contents of an image in terms a 1000-object coordinate space, and can be read in these terms. For example, the wine bottle space vector shows higher activation in the human world objects (roughly 400-950) than the natural world (0-400ish and 950-1000), these clusters are seen in the 2-confuser matrix for AlexNet trained on ImageNet in figure~\ref{fig:confuser}. Thus, a NN trained on ImageNet can be used to recognise unknown objects by comparing them to classes it knows about. For example, the cluster example given in figure~\ref{fig:ssd_flowers_4} works as although the flowers are unfamiliar, the NN labels them all as daisies as that is the closest thing it know about, that then allows the CF-NN to work with these objects. ClusterFlow is designed to work in this class vector space and relate a geometric meaning to distance in this space. This works really well with NNs due to their propensity to arrange latent variables in a real world like space. 

\subsection{Memory}

%#### Types of memory in this experiment 

Another way to analyse these experiments is through the lens of what memory the system has access to and how we might relate it to human psychology. 

\paragraph{Early visual system: ImageNet image recognition} The NN have been pre-trained and within their weights is the memory (to some extent) of the images they were trained on (ImageNet) and `how to see', which we can roughly equating with the infants learned ability to use its eyes.

\paragraph{`World' knowledge: Trained memory of what has been seen today} If we cluster the ImageNet data, then that is also the `world' knowledge. If we do not cluster with that data, but instead input a different set of data and cluster on that, then that is the world knowledge.

\paragraph{Working memory: Batch} Working memory is what is being compared or dealt with at once. 

\subsection{Relational reasoning}

%Relational reasoning tasks can be animony, symomny etc - see ppt

%The deck consists of 81 unique cards that vary in four features across three possibilities for each kind of feature: number of shapes (one, two, or three), shape (diamond, squiggle, oval), shading (solid, striped, or open), and color (red, green, or purple)
Relational reasoning has been suggested as the basis of all intelligent thought. A simple form of relational reasoning is the Set game for children that has also been used in cognitive science. In this game children are given a series of cards which can vary across three options on four axes: colour, shape, shading and number, for example: 4 red squares, 3 blue triangles, 1 yellow star, and the task is to match cards on one of the axes. There are two tasks: set and antiset. A set must match on one axes, for example \{one yellow star, one blue triangle, one red square\} matches on number. An antiset must match on no axes, as in the example given earlier in this paragraph. The children are not told which axes to match on. This measures how well the children have learned to understand relations between objects and flexible categorisation. Working memory is involved as well, as the children must compare the cards that they have. In this paper, we have set this task up for an A.I. to demonstrate that ClusterFlow adds the capability for relational reasoning to CNNs, and we have also added a task: `same, same, different' (SSD) where three images are presented and the AI must classify the set as ssd and pick the odd one out.

\subsection{Cat/dog identification task}

In \cite{quinn1993evidence} infants were familiarized with a series of pairs of familiar stimuli (i.e. two cats), and for the test they were shown either two familiar stimuli (i.e. cat-cat) or one familiar and one novel stimuli (cat-dog), and the amount of time the infant spent looking at both stimuli is recorded \cite{quinn1993evidence}. The infant is updating its internal representation of what it expects of the world and maintains fixation on the novel stimuli to do this, hence this updating is related to the looking time\cite{mareschal2000connectionist}. Note that the internal representation of recent events is a type of visual memory, so there is memory in these systems. A preference score\cite{quinn1993evidence} for the novel image is given by:

$$
\mathrm{preference} (\%) = 100\times\frac{\mathrm{time}_{\mathrm{novel}}}{\mathrm{time}_{\mathrm{novel}}+\mathrm{time}_{\mathrm{familiar}}}
$$

If the novel stimuli is believed to be a member of the familiar category, it is not any more interesting than the familiar stimuli shown at the same time, so the infant looks at both equally (a preference score of 50\%). If the novel stimuli is believed to be a member of a different category, the infant will look more at the novel stimuli. With the cats and dogs experiment it was found that novel dogs were preferred, so a dog cannot be a type of cat, but as novel cats were not preferred, a cat can be a type of dog, i.e. $cat \in dog$, $dog \notin cat$. This is an asymmetry in category identification. We measure this asymmetry as:

$$
\mathrm{asymmetry}\: \mathrm{measure} (\%) = \mathrm{preference}_{dog} - \mathrm{preference}_{cat}
$$

so positive numbers indicate an asymmetry in the experiment towards dogs and a negative number indicates an asymmetry towards cats (and the precise measure of preference varies with experiment).

The cause is related to the features of cats and dogs in the photos, i.e. the image statistics (which comes from the real world statistics of the species). In \cite{quinn1993evidence} they measured the distances of various features (nose length, nose width etc) in the pictures and found that the dog categories had a larger range than the cats. In \cite{mareschal2000connectionist} they find that the cat data in within 2 standard deviations of the dog means on many dimensions to the point where around 50\% of cats could conceivably be classified as dogs based on the numbers, but only 11\% of dogs could be classified as cats. The classes needed to overlap (although not on all dimensions) and one class needed to subsume the other (on some dimensions) in order for an asymmetry to appear. In \cite{quinn1993evidence} they removed the asymmetry and made novel cats following dogs interesting (i.e. believed to be a new category) by showing infants dogs with features that were more similar (i.e. a smaller range and I imagine non-overlapping with the cat set). 

In the early 2000s\cite{mareschal2000connectionist} and \cite{mareschal2002asymmetric} the authors show that autoencoders trained on the measured features of the images (this was pre-deep CNN which could do image classification) give the same answers as infants, if the loss (sum of squared error) of the networks was taken as relating to looking time. In \cite{mareschal2000connectionist} they expand the question (from \cite{quinn1993evidence}) to see if the asymmetry persists if the infant is trained on $\frac{2}{3}$ familiar:$\frac{1}{3}$ novel: it does. In \cite{mareschal2000connectionist} they discuss catastrophic forgetting due to retrospective interference in attempting to answer the question whether learning one category then another will make it easier to learn the second, as there is something to compare the novel exemplars to 
%[[Eimas, Quinn, and Cowan, (1994)]] 
or whether the second category will interfere with the first (retrospectively, so backwards in time). For both autoencoders and infants if the second category has a large feature overlap with the first (i.e. first = cats, second = dogs) there is retrospective interference and forgetting of the first category (as measured by cats being interesting again as if they had not been seen recently), and conversely, if the second category does not significantly overlap the first (i.e. first = dogs, second = cats), both infants and autoencoders do not forget the first category, i.e., we see asymmetry in these experiments as well. These were done with $\frac{3}{5}$ familiar - test 1 - $\frac{2}{5}$ novel - test 2 and error and looking on test1 and test2 was compared.

%[[NTS]] Aside: I am sure that there was some work on playing a 'meow' sound alongside the cat photos and a 'woof' sound alongside the dog photos which removes the infant's confusion. This is essentially just providing a aural label to the images (although one derived from animal noises rather than human words) and this turns the task into a supervised learning problem which is inherently different. I don't think I need to prove that CF will no longer make the 'mistake' if we give it correct labels.

As a standard problem in computer vision is the binary classification of the `dogs vs cats' dataset on kaggle, we will use this data and the ClusterFlow framework to investigate whether modern deep conv-NN has learned similar internal representations to babies. ClusterFlow (CF) allows us to give a geometric meaning to category membership, and CF includes both memory (working memory in terms of batch, lifetime memory in terms of hyperspatial maps) and sequence or time sensitivity (the order in which things are done matters). The question is, by adding CF to pre-trained NN can we make a system that behaves more like humans do? Note that CF is a deterministic semi-supervised algorithm put on top of pre-trained NNs. It is a difficult thing to decide whether the addition of CF counts as re-training or not. It is re-training in that CF requires some data to use to draw its hyperspatial maps which are then used, however it is not retraining to the task as these maps are not drawn in response to solving the question (as would happen in standard retraining) but deterministically based on what was previously learned by the deep-NN. I compare these results to standard retraining methods where the top layers of the pre-trained NN are retrained on the cat/dog classification task (which is not the task that the babies are doing). Thus, we add CF to CNNs to see if this allows human-like behaviour on these simple tests.

%\paragraph{References:}

%[[86:]] Quinn, P. C., Eimas, P. D., & Rosenkrantz, S. L. (1993). Evidence for representations of perceptually similar natural categories by 3-month-old and 4-monthold infants, Perception, 22, 463-475

%[[87:]] A connectionist account of asymmetric category learning in early infancy, Mareschal et al, 2000

%[[88.]] Asymmetric interference in 3- to 4- month olds sequential category learning, Mareschal et al 2002

\subsubsection{Missing data in chemistry datasets}

Although this paper is primarily concerned with using CF as a layer to a pretrained NN, the clustering algorithm can be used on any data, and can cope with missing datapoints. As this is a problem frequently found in chemistry, we test CF on a chemistry dataset with missing data. The dataset records various physicohemical properties relating to how a molecule performs as a solvent and the molecules are labelled with functional groups, so each molecule has many labels. The task is to cluster the solvents based on the solvent data and see how well that correlates to functional group labels.

\section{Methodology}

In this section we will go through how ClusterFlow was built and works, before laying out the experimental methodology.

\subsection{Designing ClusterFlow}

The equation below relates the hypothesis function for ClusterFlow, $F_{CF}$, and its inputs, the class space vectors, $\vec{y}(\Omega_c)$, the label(s), $\hat{y}(\Omega_l)$ and the metric we use to build the clusters, $L^p$ with the output cluster $C$:

\begin{equation}
     C = F_{CF}(\vec{y}(\Omega_c), 
     \hat{y}(\Omega_{l}), L^p)
    \label{eq:cluster_flow}
\end{equation}

Note that, ClusterFlow can be used with any data, it does not have to be data from a NN, so $\vec{y}$ could be a vector in any metric space, (with $L^p$ as the metric, of course) and $\hat{y}(\Omega_l)$ being the labels for the datapoints.

\subsection{Building ClusterFlow}

In this section we go through how ClusterFlow is put together. The full source code will be published on GitHub upon publication.

\subsubsection{Assumptions}
We make two main assumptions - that points near each other in activation space (or some subset of activation space) - see Terms below - are likely to share similar labels (and that the size of that distance has some bearing on the likelihood); and that the input data will need to be parsed in batches (i.e. we are dealing with `big data').

\subsubsection{Terms}
To explain the code, we first need to define a few terms.
\begin{description}
\item[Activation]
As ClusterFlow was originally developed to map the activations of hidden layers of a CNN and relate those activations to image classes. Thus the activation table is set up with `neurons' being the width of the activations and `images' being the datapoints.

Activations are the feature vectors, $y(\Omega_c)$, a filename and either a label or list of labels. which is saved in a hdf5 file that the neurons refers to the activations. An activation $a$ is given below, where Act is the class activation. 

$$
a = \mathrm{Act}(y(\Omega_c), \{\hat{y}(\Omega_l\})
$$
\item[Activation Space]
The hyperspace of the same dimension as the feature vectors (i.e. if we are working with 23 features in our input, then the activation space is 23 dimensional.
\item[Label]
A label is one of the classes that an activation may be associated with (e.g. `daisy').
\item[Cluster]
A cluster is a region of activation space (or a sub-space of activation space) defined by some form of bounding box and a mapping of those activations and labels that we have seen that fit within it. Clusters may also contain other clusters.
\item[Mixed and Pure clusters]
A mixed cluster is one such that there is no one label in common between all of the activations within it.

A pure cluster is one that such that all the activations within it share at least one label. 
\item[Bounding Box]
Each cluster is defined in space by an axis-aligned hyper-cuboid. This box is defined by two vectors that describe two opposed corners of the box.  
\item[Lower Dimensional Bounding Box]
When dealing with activations coming from hidden layers in Neural Networks, most activations will be zero\cite{GaleObjectDetectors}, so often we will be dealing with sparsely populated activation vectors. To avoid giving incorrect weight to these zeros, the lower dimensional bounding box only pays attention to non-zero values in the activations that comprise it. For example, for if an activation, $\vec{a}=[0, 0.1, 3, 0, 5]$, is added to the box, it will only add the non-zero dimensions, $a_1=0.1$, $a_2=3.0$ and $a_5=5.0$.

Thus boxes (and by extension clusters) can exist in a lower dimensional `shadow' of activation space. This makes sense as not all dimensions of a given activation will be expected to contribute to determining the output class, and indeed with NNs we are expecting a lower dimensional manifold that our data lives on and we wish to prevent hacking via escape dimensions. 

\item[Partial Dimensional Bounding Box]
Some datasets will not have measurements for all values for all dimensions of activation space. The partially dimensional bounding box allows for input activations to have `None' values for some fields, and has additional functions to allow for the uncertainty of that value in determining what is in (or not) that box. 

Basically, if your input dataset does have missing values, use this, otherwise stick to LowerDimensionalBoundingBox.

\end{description}

\subsubsection{The ClusterFlow Algorithm}
ClusterFlow works by recursively building up a series of hypercuboid clusters in activation space, then refining these clusters into a tree structure such that the leaf nodes are all pure clusters.

\begin{figure}[htp]
    \centering
    \includegraphics[width=0.6\textwidth]{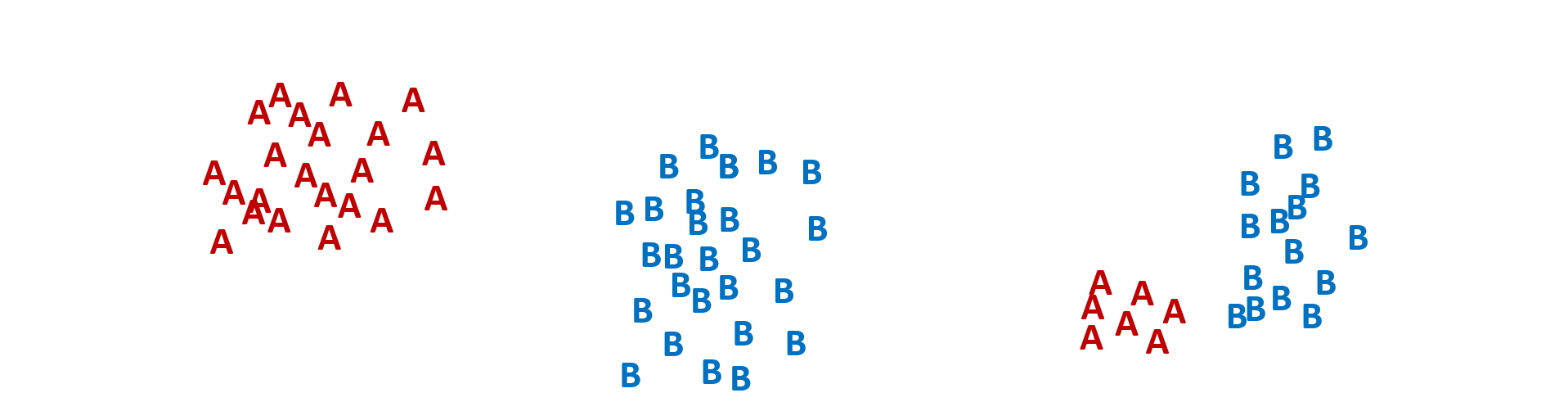}\\
    a. Starting data\\
    %\hline
    \includegraphics[width=0.6\textwidth]{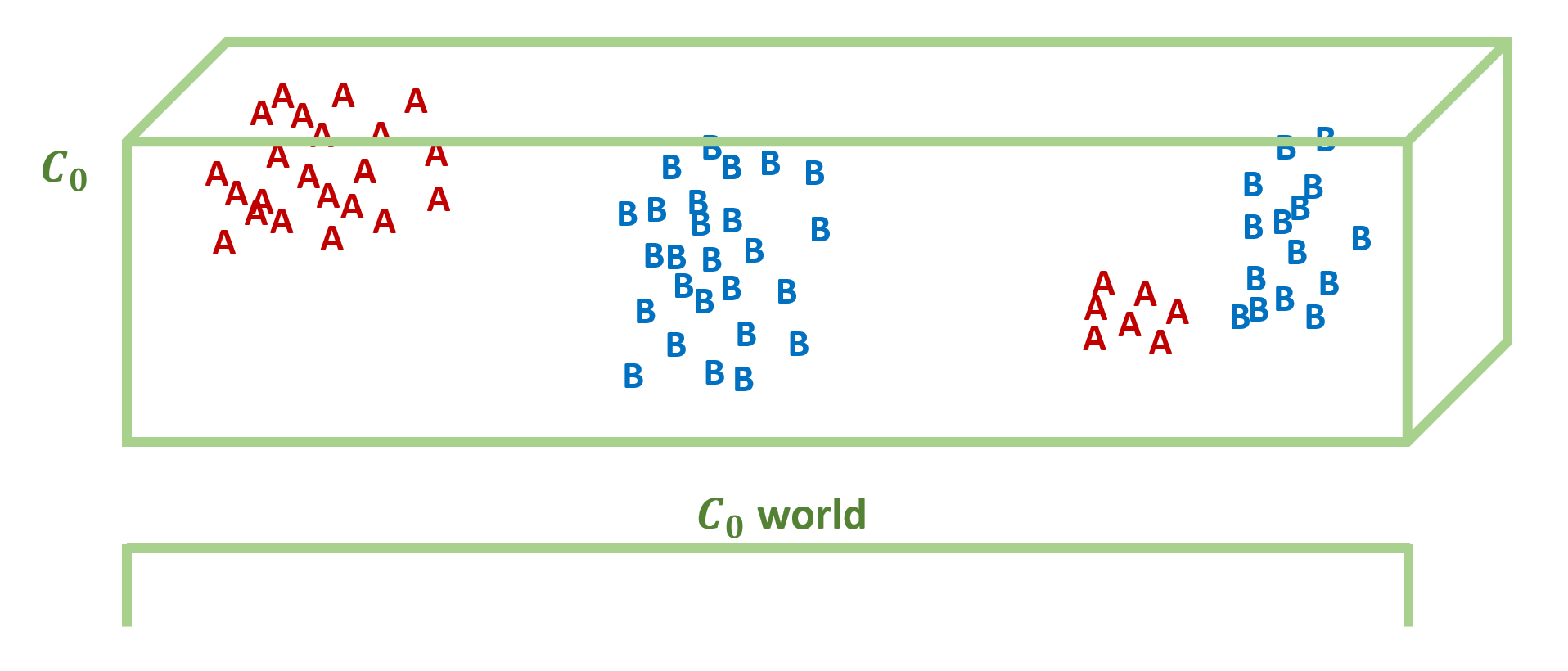}\\
    b. Encapsulate world knowledge with top level $C_0$ cluster\\
    %\hline
    
    \includegraphics[width=0.6\textwidth]{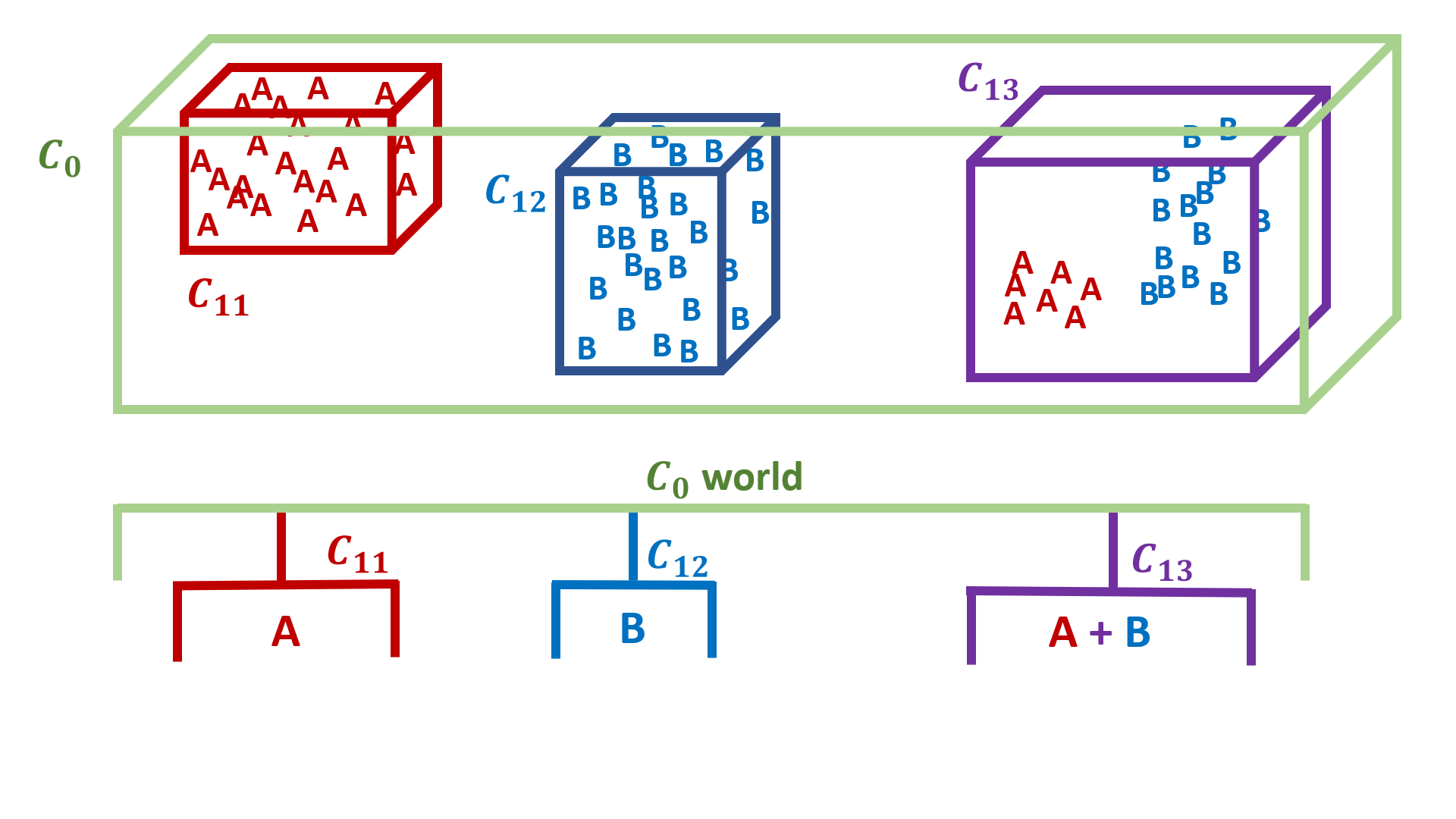}\\
    c. Recurse into top level cluster looking for pure clusters\\
    %\hline
    \includegraphics[width=0.6\textwidth]{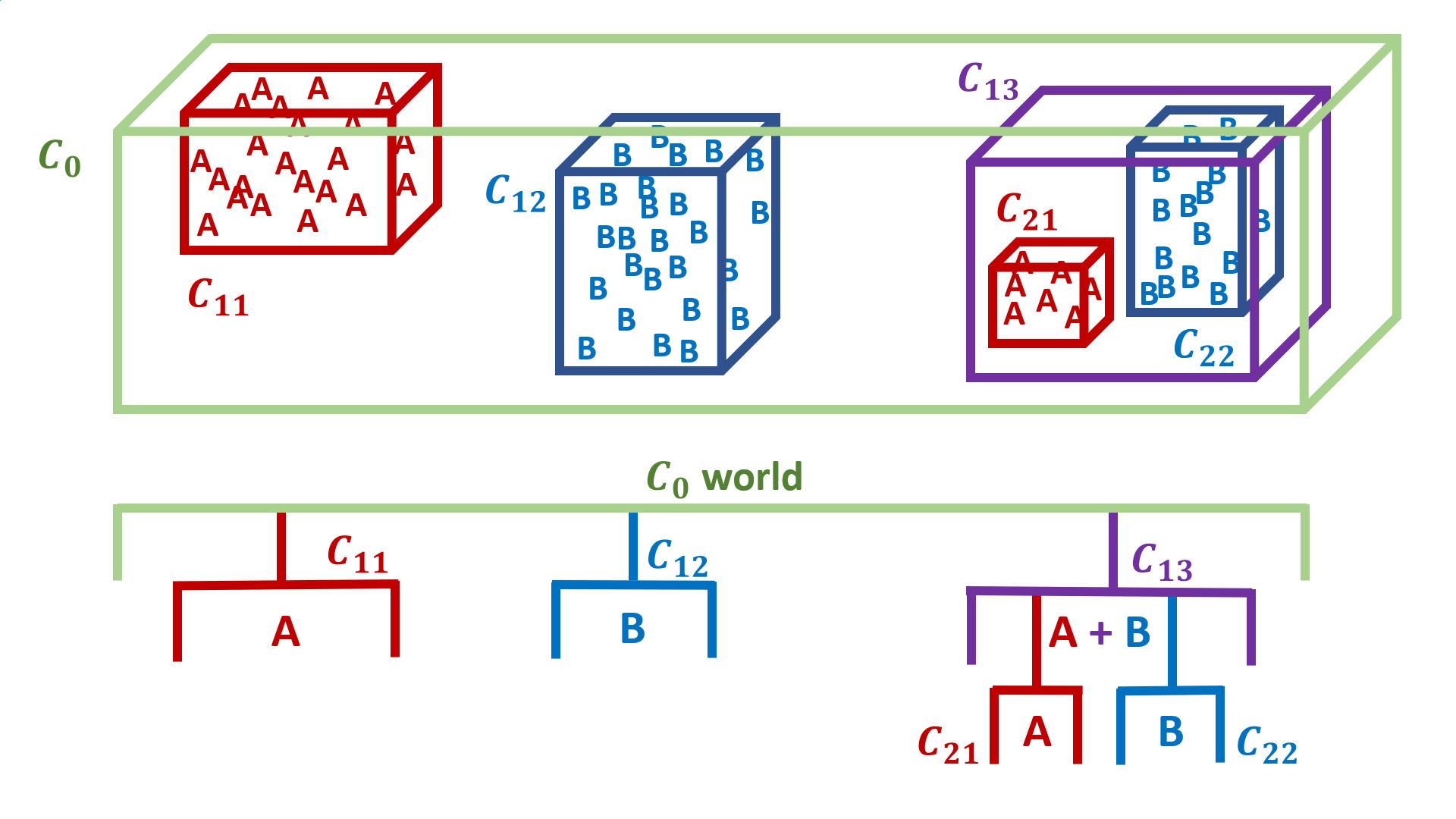}\\
    d. Recurse further into mixed clusters until you get pure clusters\\
    %\hline
    \caption{Recursive clustering algorithm.}
    \label{fig:cluster_alg}
\end{figure}

As the algorithm is primarily recursive.
First we create a set of clusters to start with. For this you can use any standard clustering algorithm - We chose detK but k-means and k-means++\cite{32} are also included. You take as many points as possible of each class (all of them ideally) and generate the level 1 cluster(s) for each class separately. For example, in ImageNet you'd run the 1,300 tinca tincas, then run the 1,300 tiger sharks, etc. We then combine all of these clusters into a single set that delineates the world data. For example, in figure~\ref{fig:cluster_alg}a we see that there are three clusters, we then cluster A into two clusters and B into two clusters, and $C_0$, the world cluster, is slightly bigger than those cluster. 

Next we allocate the points to each cluster (note: regardless of which class created it), and expand overlapping clusters. In figure~\ref{fig:cluster_alg}b the points are allocated to the 4 found clusters: $C_{11}$, $C_{12}$ and the two overlapping clusters of A and B have been merged into a mixed cluster $C_{13}$. This can (and almost certainly will have to) be done in batches.

For each cluster now if it is mixed, we recurse into it, using detK on the activations contained within it to generate a set of clusters, then mapping the activations to those clusters and recursing into any mixed children and so on.

There is a slight wrinkle in that a cluster needs at least two points of a given label to process them, so singletons get rejected at this stage, gathered up and run as a final batch at the end.

Eventually we will have a tree hierarchy of clusters which are such that every leaf node is either pure or contains at least one singleton activation that couldn't be broken down further.

An example of a generated cluster tree is given in figures~\ref{fig:solvent} and ~\ref{fig:cat_dog_cluster}. For further worked examples, see the examples directory in the code repository.

\paragraph{Expanding clusters}

When clusters are expanded, we go slightly beyond the point in each direction, so each point can bring with it a subspace. We use the Mehalanobis distance (see Appendix), which divides the distance by the standard deviation of points in that direction. For example, if we wanted to expand a box by 10\% and $-2\leq y_1\leq2$ and $-10\leq y_1\leq10$ then we would expand $y_1$ by 0.4 and $y_2$ by 2, so the same distance means different things in different dimensions. This allows us to map the manifold, and is the natural metric to use for exploring the manifold.

%¬!!![[put in picure mehalanobis]]

\subsubsection{Using clusters to guess labels for activations}

\begin{figure}[htp]
    \centering
    \includegraphics[width=0.9\textwidth]{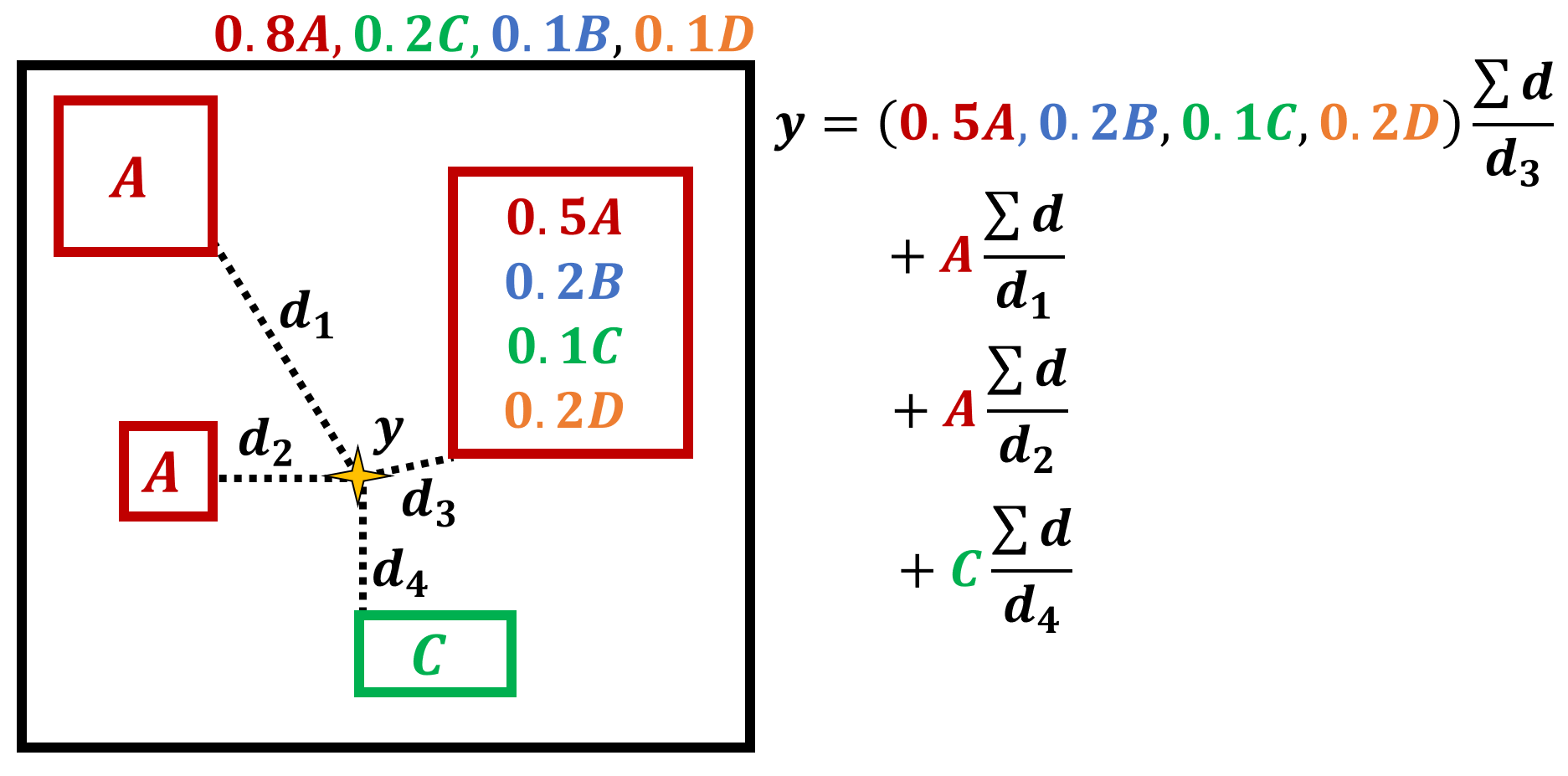}
    \caption{Assigning guesses to points}
    \label{fig:assignment}
\end{figure}

Once the clusters have been built we can use the cluster `maps' in $\Omega_c$ to assign labels to input class vectors, $y(\Omega_c)$.  If $y$ is inside a pure cluster, it gets the label of that cluster, if it is inside a mixed cluster, it would get the labels in order of majority (i.e. top-1 would be the majority label, the second label would be the next most common label). If it is not in a cluster then it is assigned guesses based on nearby clusters, an example of how this is done is shown in figure~\ref{fig:assignment}. The distances, $d$ between point $y$ and the nearest boundary of the nearest clusters are measured, and the classes guesses are calculated weighted by the closeness of the clusters, the equation is shown in figure~\ref{fig:assignment}. The distance metrics used can be $L^0$, $L^1$, $L^2$ or Mehalanobis, and the computation can be done in subspaces in the case of missing data, zero data or subspace hypercuboids. 

\section{Testing ClusterFlow}

\subsection{Experiment 1: Missing data: Clustering on a chemistry dataset}

ClusterFlow is a hierarchical clustering algorithm that can deal with missing data. To demonstrate this, we used a dataset from chemistry, as the problem of missing and incomplete data is common in experimental sciences. The dataset was a solvent database\cite{murray2016application} containing 634 solvents with 22 parameters e.g. boiling point, molar volume, surface tension, Catalan, Laurence and Hansen’s solvent parameters, molecular weight. Only 40 solvents have data for each parameter. Two more databases were created with the same molecules. A chemical group database with 167 labels that refer to chemical functional groups, e.g. alkane, alcohol, ketone, sulfide, benzene, dioloane and a structure database with three parameters: SMILES, InChI, elemental makeup. The task is to cluster the data based on the solvent properties using the functional groups as labels. The control is to use standard k-means (det-k) to cluster the full dataset. 
% NTS - the lathanides dataset would go well with this
% NTS 2 - comparing to the clusters after a NN had learned this data would also be useful.

%Measure asymmetry in `surprise'. 

\subsection{Experiment 2: Resilience to fooling images}

Fooling images came from~\cite{40} (N=5000). We compare pre-trained AlexNet (as they did in~\cite{40}) with pre-trained AlexNet with ClusterFlow trained on top using the ImageNet data (world knowledge cluster). AlexNet is not retrained. 

\subsection{Experiment 3: Relational learning}

\subsubsection{Setting up the datasets}

\begin{table}
    \centering
    \begin{tabular}{|c|c|c|c|c|}
        \hline
        Dataset & \multicolumn{2}{|c|}{Within class} & \multicolumn{2}{|c|}{between class} \\
        & \multicolumn{2}{|c|}{a:a}& \multicolumn{2}{|c|}{a:b} \\
        \hline
        Imagenet & \multicolumn{4}{c|}{} \\
        \hline
        \includegraphics[width=2cm]{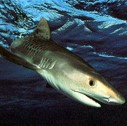} \includegraphics[width=2cm]{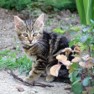} \includegraphics[width=2cm]{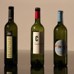} \includegraphics[width=2cm]{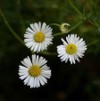}& $<L^1> = 2140$ & $<L^2> = 87$ & $<L^1>=3361$ & $<L^2> = 141$\\
                \hline
        Flowers\_4 & \multicolumn{4}{c|}{} \\
        \hline
        \includegraphics[width=2cm]{daisy_4.jpg} \includegraphics[width=2cm]{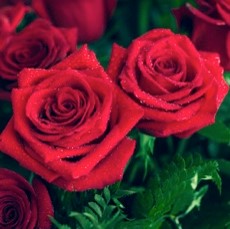} \includegraphics[width=2cm]{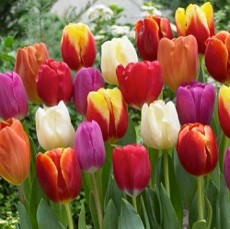} \includegraphics[width=2cm]{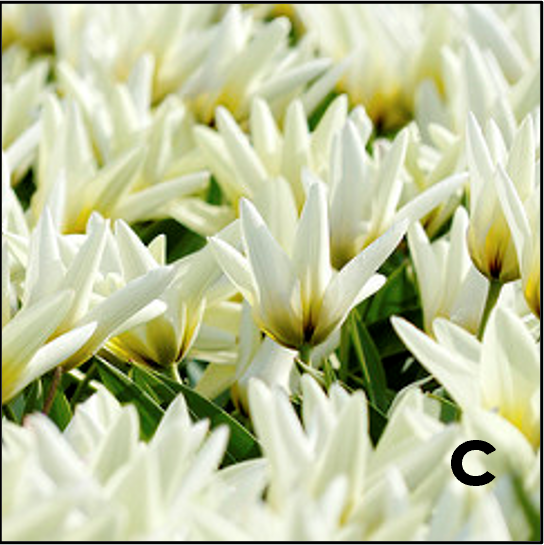}& $<L^1> = 2126$ & $<L^2> = 86$ & $<L^1>=2428$ & $<L^2> = 98$\\
                \hline
        SketchNet\_4 & \multicolumn{4}{c|}{} \\
        \hline
        \includegraphics[width=2cm]{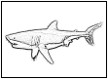} \includegraphics[width=2cm]{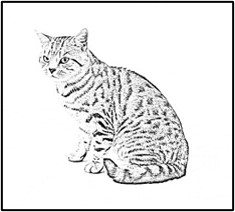} \includegraphics[height=2cm]{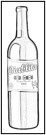} \includegraphics[width=2cm]{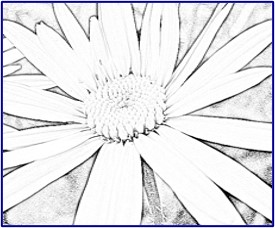}& $<L^1> =1485$ & $<L^2> = 60$ & $<L^1>=1863$ & $<L^2> = 75$\\
        \hline
        
    \end{tabular}
    \caption{Datasets for the relational reasoning task.}
    \label{tab:datasets}
\end{table}

Testing was done with singly, labelled data for example $\{ x, ``a"\}$ where $x$ is a vector and $``a"$ is a label. ImageNet\_4 dataset was made by choosing four classes from across ImageNet: Tiger sharks (3), tabby cats (281), daisies(985), wine bottles (907). $\vec{y}(\Omega_c)$ came from putting the original photos through pre-trained AlexNet. To get the labels, $y(\Omega_l)$, we used a CF layer that had been trained on the whole of ImageNet (the world knowledge cluster) to guess up to five labels per point. Flowers\_4 includes novel exemplars of the `daisy' class, which is known to AlexNet, and three new-to-AlexNet flower classes of `dandilions', `roses' and `tulips'. Labels were assigned by using CF to guess up to 5 labels. SketchNet\_4: The four classes from ImageNet\_4 converted to sketches using a GIMP script, and labels were guessed as for ImageNet\_4 and Flowers\_4.  
%[[add to SI]].

%To guess multiple labels: 
%load world cluster (teh world knowledge from alexNEt)
%laod 3 datapoints
%guess 5 labels for each point
%attempt to clyuster
%output ios cluster obkect 

\subsection{Experiment 4. Irreflexivity in cat/dog identification task}

\subsubsection{Experiment 4.1.} First we give ClusterFlow all the cats from the training set (9280), create a cluster, see how many test dogs (1120) are outside that cluster, this is a measure of surprise (as the network has `believes' the dogs outside the cluster are something different to all the cats seen before) on a novel unfamiliar stimuli. Repeat with a cluster of dogs (training set), and test cats. To get the response of NN-CF to a novel familiar stimuli, use the test cats on the cat cluster and testset dogs on the dog cluster. Pets have single, known labels (i.e. `cat', `dog'). 

\subsubsection{Experiment 4.1. Controls}

\paragraph{Test with training set} Use the unfamiliar pets from the training set. This controls for the size of the test set as the training set is the same size. 

\paragraph{Clustering with all the cats and dogs at once} Taking the entire train set of images with both labels, we made a large cluster and looked at the cluster structure to see how many pets were assigned to `pure' clusters (a measure of separability of the classes in this space) to see if there was any asymmetry. The cluster has a 100\% accuracy. 
%[[There was no asymmetry, demonstrating that the ordering is necessary for the asymmetry effect]].

\subsubsection{Experiment 4.2. Changing category feature ranges} In the literature, the asymmetry effect goes away when a set of dogs with a smaller range of features was used, so to compare, we chose to use 9 breeds consisting of 3 small dog breeds (Jack Russells, corgis and chihuahuas), 3 medium dog breeds (Labradors, whippets and rottweilers) and 3 large dog breeds (great dane, Saint Bernard and bloodhounds). Five of these breeds are classes in ImageNet: chihuaua, bloodhound, great dane, Saint Bernard and Labrador-retriever. 
%It was said %[[dunno if i have a reference for this]] that 
Dog breeds are more different from each other as they have been bred for different jobs. This set tries to cover as much of that space as possible. Nine clusters were built for the nine breeds, and two clusters of test dogs and test cats were also built (all clusters contained 1120 points) and the experiment 4.1 was repeated with these clusters. Means were taken of the surprise. Different dog breeds were compared (by size).

\subsubsection{Experiment 4.2. Controls}

\paragraph{Testing if the single breeds are in the original train set dog cluster} The surprise (1-accuracy) for the single dog breeds found outside the dog cluster. Finds the accuracy of the single dog breeds data. 

\paragraph{Repeating experiment 1 with 9280 dogs from the single breed datasets to verify that the 9 single breeds roughly cover the same space} We took a balanced set of randomly selected single breed dogs ((9280/9) dogs from each breed, set of randomly) and built a single large cluster from them and then repeated experiment 1 to see if the asymmetry was present in this dataset. 
%[It was]

\subsubsection{Experiment 4.3: Standard NN retraining} 

MobileNet was retrained on the kaggle cat vs dog task with both sets of pets mixed together and input in a random order.
%NTS chekc if this waws MobileNet

\section{Results}

\subsection{Experiment 1. Chemistry dataset}

\begin{figure}[htp]
    \centering
    \includegraphics[width=\textwidth]{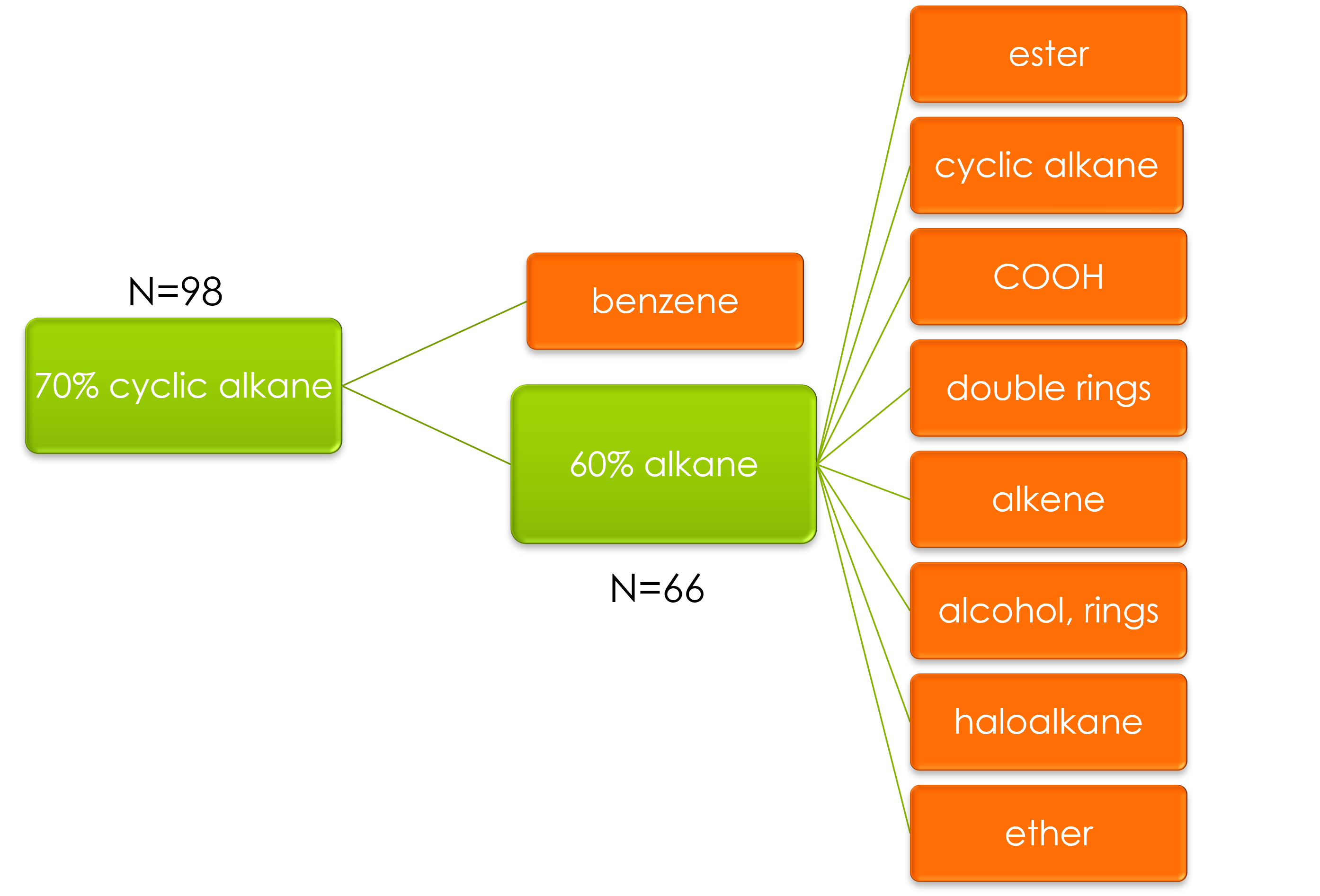}
    \caption{Part of the solvent space for the chemistry clustering experiment.}
    \label{fig:solvent}
\end{figure}

The k-means control could only cluster 22 solvents. By dropping out various two chemical groups: dielectic constant, dipole moment N = 304, d=23, and 2 clusters were found to be best at describing the data with the physical properties that cause most of the separation: Mw (molecular weight), MP (melting point), BP (boiling point), mV (molar volume). Excluded chemical groups: Mw, MP, BP N = 287,d=22, 2 is the best at describing the data, with the physical properties that cause most of the separation: dipole moment, dielectric constant, mV (molar volume). ClusterFlow was able to cluster all the data across all the solvent parameters. Chemistry data is `spiky', as the resulting cluster was very complex with 53 clusters width at the top layer and 20 deep and different overlapping clusters. Nonetheless, the cluster formations look chemically sensible, part of the cluster is shown in figure~\ref{fig:solvent}. This has demonstrated that ClusterFlow can cluster on incomplete data. A simple neural network autoencoder was also trained on this dataset, and then the latent variables were clustered using the functional group labels, and this gave a much neater and more compact cluster structure, agreeing with previous work finding that NNs reorganise data onto a lower dimensional, and sensible manifold.
%% citations!

\subsection{Experiment 2: Resilience to fooling images}

\begin{figure}[htp]
    \centering
    \begin{tabular}{|p{2cm}|p{2.5cm}|p{2.5cm}|p{2.5cm}|p{2.5cm}|p{2.5cm}|}
    \hline
       &\includegraphics[width=2.5cm]{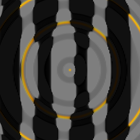}  &  \includegraphics[width=2.5cm]{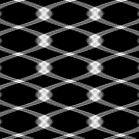} &\includegraphics[width=2.5cm]{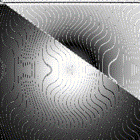} &\includegraphics[width=2.5cm]{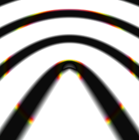} &
       \includegraphics[width=2.5cm]{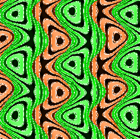} \\
       \hline
       AlexNet & 99.9\% king penguin & 
       90.1\% honeycomb &
       43.5\% ‘tusker’
        42.0\% Indian elephant & 
        99.9\% chocolate sauce &
        99.5\% monarch butterfly \\
        \hline
        AlexNet + ClusterFlow &
        0.02\% fire screen &
        0.26\% golf ball & 
        0.14\% balloon & 
        0.01\% trilobite &
        0.06\% plastic bag \\
        \hline
    \end{tabular}

    \caption{Adding a ClusterFlow layer to pre-trained AlexNet greatly decreases its confidence on fooling images, results for five example images.}
    \label{fig:fooling_zB}
\end{figure}

Using ClusterFlow as an extra layer on top of AlexNet greatly improved its resilience to hacking, demonstrating the validity of the technique and the supposition that we can map out class vector space in this manner. Figure~\ref{fig:fooling_zB} shows some examples of fooling non-images and the class assignment and confidence from AlexNet with and without a CF layer. We can see that that the confidence has dropped from $>90\%$ to <1\%. The range of confidence ratings covers the whole range, with the majority being in the 90-100\% confident bracket for AlexNet, with CF we can see that the majority confidence is below 0.05\%. Table~\ref{tab:fooling_probability} shows that 1.6\% of the fooling images where outside of the world cluster $C_0$ and were given a NAN as CF will not guess on vectors that are outside of its knowledge of the world. I.e. there were fooling images that were sufficiently different to the latent variable encoding of ImageNet by AlexNet to be geometrically outside of the manifold. This gives evidence for the sketch in figure~\ref{fig:general_idea}. As AlexNet + CF gets the same accuracy\footnote{We can retrain CF to increase accuracy by using the real labels, but CF was trained with the labels AlexNet believed to be correct}, but this combined algorithm is no longer as easily fooled, we have demonstrated how to add resilience to visual hacking attacks to NN.

\begin{table}[htp]
    \centering
    \begin{tabular}{|c|c|c|}
    \hline
        No. of activations with & AlexNet & AlexNet + ClusterFlow \\
        \hline
        $p_x > 0.95$ & 27.5\% & 0\% \\  
        $p_x > 0.90$ & 36.0\% & 0\% \\  
        $p_x > 0.50$ & 69.0\% & 0\% \\  
        $p_x > 0.20$ & 91.3\% & 0\% \\  
        $p_x = NAN $ & 0\% & 1.6\% \\ 
        \hline
    \end{tabular}
    \caption{AlexNet is overconfident on its identification of fooling images, adding ClusterFlow reduces this.}
    \label{tab:fooling_probability}
\end{table}

\begin{figure}[htp]
    \centering
    \includegraphics[width=0.45\textwidth]{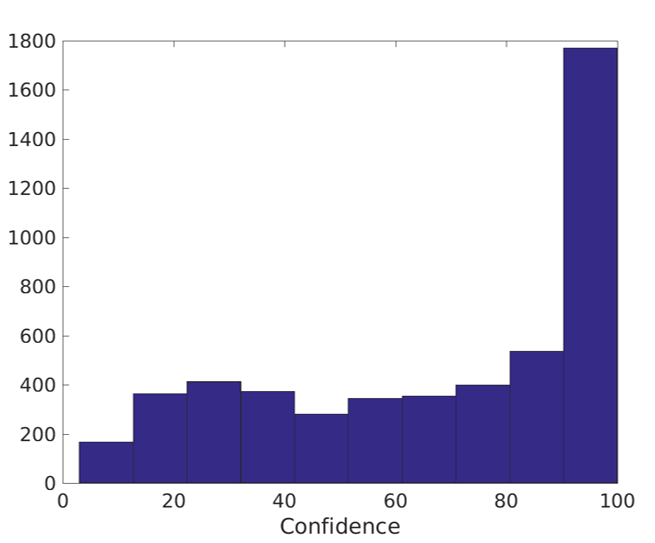}
    \includegraphics[width=0.45\textwidth]{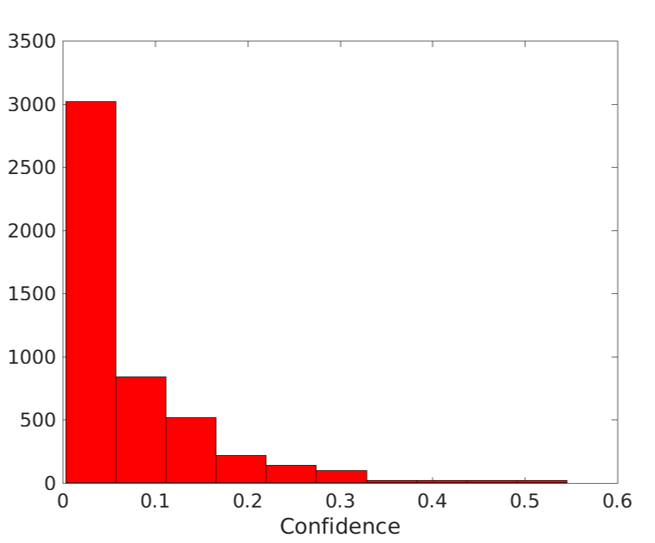}
    \caption{The confidence rating on the fooling images from AlexNet (left, blue) and AlexNet with ClusterFlow (right, red). These images should have assignments with very low confidence as they do not correspond to anything in ImageNet.}
    \label{fig:overconfidence}
\end{figure}

\clearpage

\subsection{Experiment 3. Relational reasoning \label{r:relreas}}

The implementation of the relational reasoning task. For the set task: 
$$
C_{set} = \{(x, ``a"),  (y, ``a"), (z, ``a")\}
$$
we get $C_0$ is pure and number of rejects = 0, for the SSD task:
$$
C_{ssd} = \{(x, ``a"),  (y, ``a"), (z, ``b")\}
$$
we get $C_0$ is mixed, $C_1$ is pure and contains 2 points, and number of rejects = 1, and 
$$
C_{antiset} = \{(x, ``a"),  (y, ``b"), (z, ``c")\}
$$
we get $C_0$ is mixed and number of rejects = 3. These examples are part of the unit tests and as table~\ref{tab:real_reas_results} shows, CF got 100\% on these, demonstrating that the method of using labeled cluster to perform the set, antiset and SSD relational reasoning tasks is sound.   

\begin{figure}
    \centering
    \includegraphics[width=5cm]{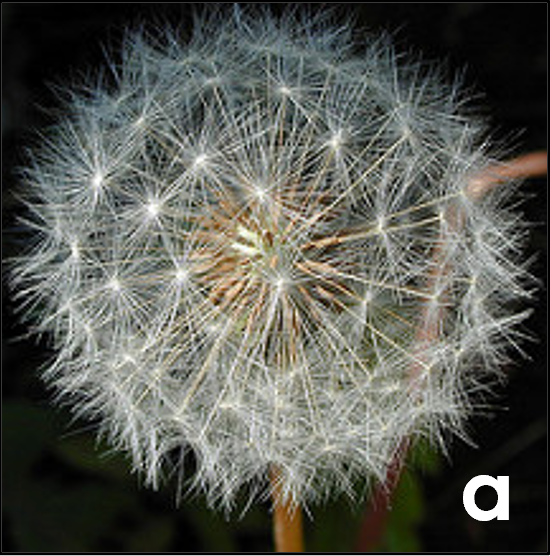}
    \includegraphics[width=5cm]{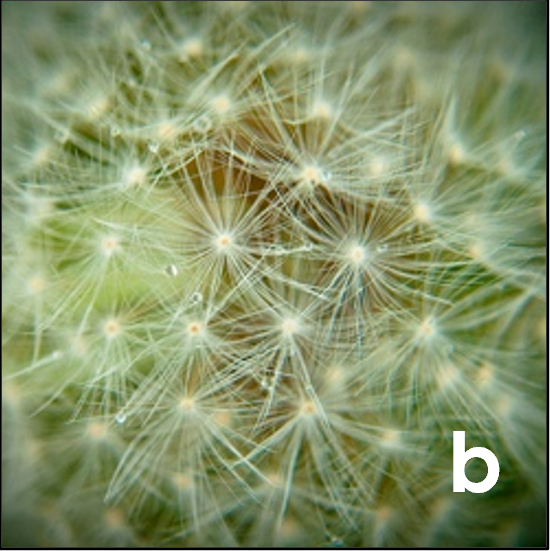}
    \includegraphics[width=5cm]{dandi_3.png}
    \caption{The SSD task with three flower examples from the flowers\_4 dataset}
    \label{fig:ssd_flowers_4}
\end{figure}

Figure~\ref{fig:ssd_flowers_4} shows an example for the SSD task. There are several dimensions these images could be compared on. A human might say that they are all the same as they are all pictures of flowers, or all the same as they contain white as the main colour. Yet, a human would have no difficulty in picking the odd one out, c, as it is a different species of flowers to the other two. The final top level cluster for the 3 flowers was \code{\{`n01914609'\}} or `jellyfish'. Examples $a$ and $b$ share three labels \code{\{`n01914609', `n07730033', `n02319095'\}}, $b$ and $c$ share 1 label, $a$ and $c$ share two labels \code{\{`n01914609', `n11939491' (daisy)\}}. So, $C_0$ is a pure cluster, there are no rejects and $\{a,b\}$ are more similar to each other than $\{b,c\}$ or $\{a,c\}$. Note that, the top class was not `daisy' as they are all flowers, but `jellyfish', perhaps due to the texture and shapes. 

Table~\ref{tab:real_reas_results} shows the results for relational reasoning using AlexNet. For ImageNet\_4, which is made up of classes and exemplars it has seen before, AN+CF is about to get around 20-33\% correct on each task. Given that AlexNet only has a top-1 accuracy of 50-60\% for ImageNet this is pretty good. For the Flowers\_4 dataset, it is able to get between 14-50\% on the task. Three of the flowers classes in the datasets are novel, so the fact that it can get any correct at all shows that AN+CF is able to do relational reasoning on novel objects. The labels chosen for the flowers depends on the world knowledge of AlexNet, so the high score for anti-set suggests that AlexNet `sees' the different flower types as rather different from each other. It also exceeds the accuracy in identification (as daisies have been seen, the accuracy is less than 25\%) demonstrating that AN+CF can compare novel objects. Similarly, SketchNet removes a lot of the textures that CNNs rely on to assign classes to images, so these inputs look novel (and are closer to each other, see table~\ref{tab:datasets}), and the AN+CF can do the task at some low level. Note that as all these image are line drawings, they do look more similar and this is a very difficult problem for AN. 

\begin{table}
    \centering
    \begin{tabular}{|c|c|c|c|c|}
    \hline
        Dataset & Set & SSD & Antiset & Accuracy  \\
        \hline
        Unit tests & 100\% & 100\% & 100\% & N/A \\
        ImageNet 4 & 20\% & 25\% & 33\% & 50-60\% (top 1)\\
        % NTS actually find out the accuracy for that
        Flowers 4 & 14\% & 31\% & 51 \% &23.8\% \\
        SketchNet 4 & 2\% & 10\% & 21 \% & 5.2\% \\
        \hline
    \end{tabular}
    \caption{Results for relational reasoning with AlexNet trained on ImageNet. Accuracy results are the top-1 accuracy for those classes in AlexNet.}
    \label{tab:real_reas_results}
\end{table}

\subsection{Experiment 4. Irreflexivity \label{res:catdog}}

\begin{figure*}[t!]
    \centering
    \begin{subfigure}{0.5\textwidth}
        \centering
        \includegraphics[height=2.5in,width=\textwidth]{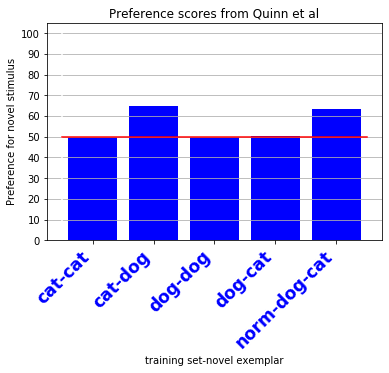}
        \caption{Preferential looking scores in 3-4month old infants\cite{quinn1993evidence}}
    \end{subfigure}%
    ~ 
    \begin{subfigure}{0.5\textwidth}
        \centering
        \includegraphics[height=2.4in,width=\textwidth]{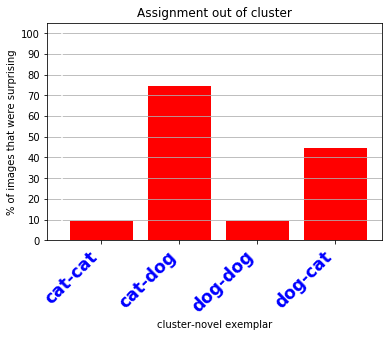}
        \caption{Experiment 4.1: Preferential inaccuracy for familiar and novel test exemplars shown to familiar-class-only cluster built on top of MobileNet}
    \end{subfigure} \\
        \centering
    \begin{subfigure}{0.5\textwidth}
        \centering
        \includegraphics[height=2.5in,width=\textwidth]{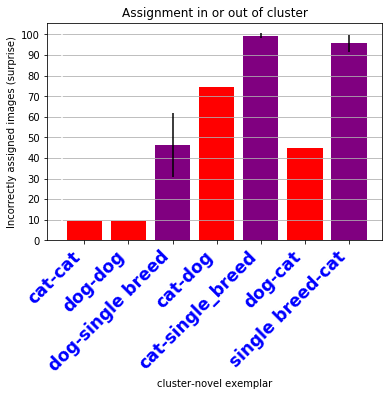}
        \caption{Surprise in Exp. 4.2. Using single dog breeds to reduce the variance of the dog clusters}
    \end{subfigure}%
    ~ 
    \begin{subfigure}{0.5\textwidth}
        \centering
        \includegraphics[height=2.4in,width=\textwidth]{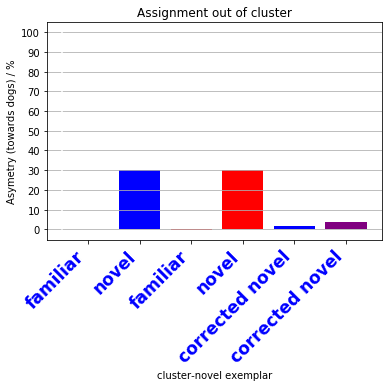}
        \caption{Asymmetry measures towards dogs. Blue: human infant data\cite{quinn1993evidence} red: NN+CF on cat vs dog dataset; purple: NN+CF on corrected single breed dog dataset}
    \end{subfigure} \\
            \centering
   % \begin{subfigure}[t]{0.5\textwidth}
    %    \centering
     %   \includegraphics[height=2.5in,width=\textwidth]{MobileNetSurpriseExp2.png}
    %    \caption{Placeholder for a control - }
    %\end{subfigure}%
    %~
    \begin{subfigure}{0.5\textwidth}
        \centering
        \includegraphics[height=2.4in,width=\textwidth]{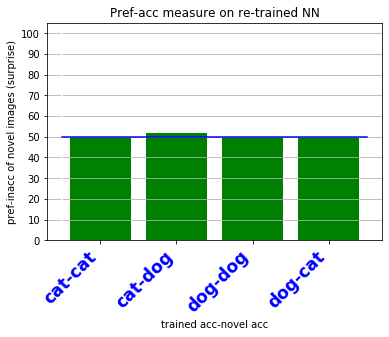}
        \caption{Preferential inaccuracy from retraining MobileNet on the cat/dog identification task}
    \end{subfigure} \\

    \caption{Example results from MobileNet.}
    \label{fig:MobileNet}
\end{figure*}

Example data is shown for the MobileNet CNN in figure~\ref{fig:MobileNet}. We see that the preferential inaccuracy is in the same direction and roughly a similar amount in the standard experiment (4.1), figure~\ref{fig:MobileNet}b, demonstrating that MobileNet + CF makes the same `mistake' as a human infant. This was the experiment were the cats were first clustered and then we see how many dogs are outside of that cluster to get the cat-dog surprise measure. This shows that MobileNet's internal latent variable representation of the world $y(\Omega_c)$ is set up similarly to that of a human infant. In the control experiment where the pets were entered at the same time the asymmetry effect disappeared, demonstrating that ordering is part of the cause of the effect, see figure~\ref{fig:MobileNet}e.

In experiment 4.2, the dog category was replaced with a single dog breed and we see that the surprise from cat to single breed dog and single breed dog to sat are much higher due to the smaller size of the single breed cluster than the dog cluster, see figure~\ref{fig:MobileNet}c. 

\begin{figure}
    \centering
    \includegraphics[width=\textwidth]{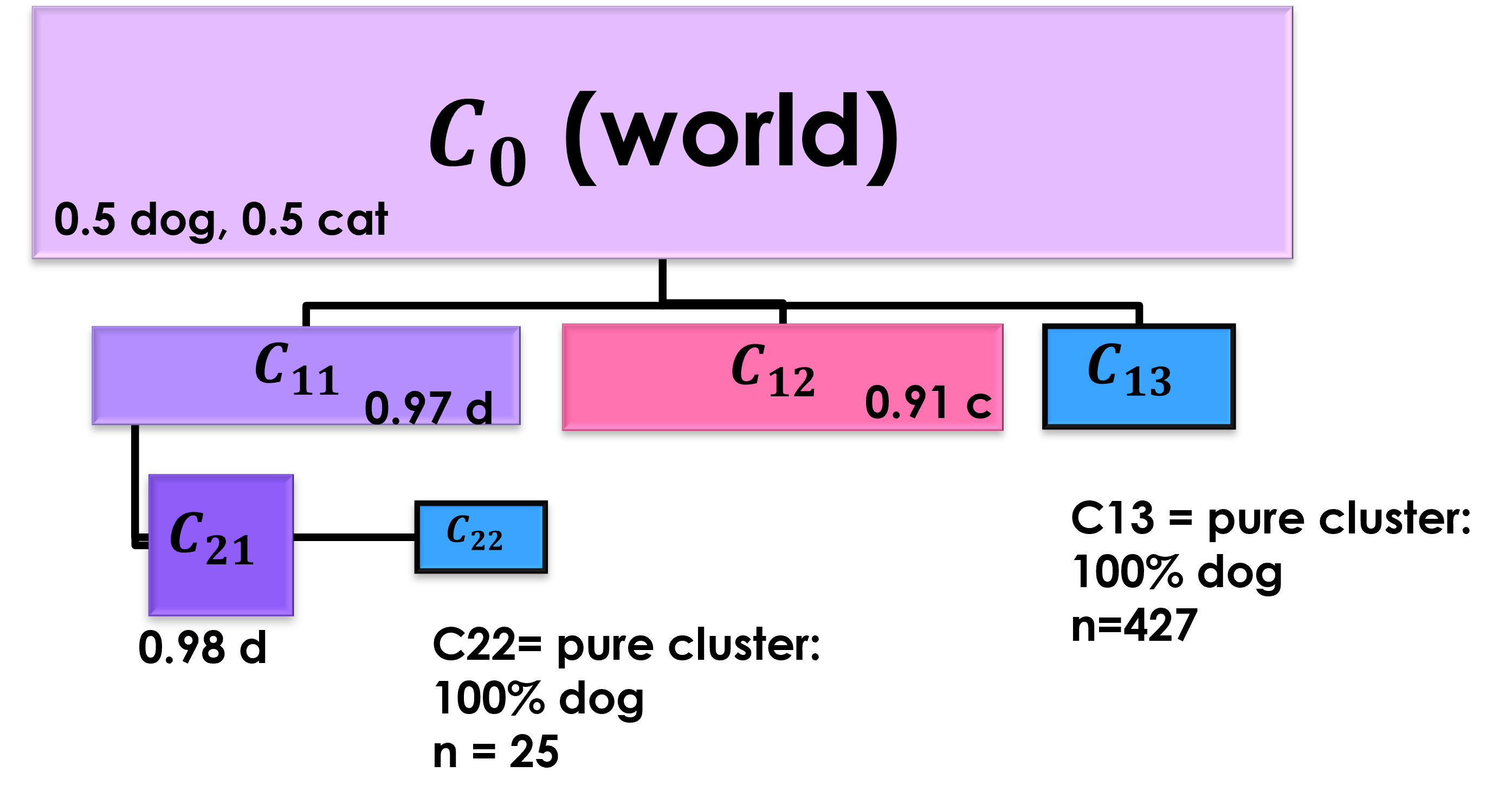}
    \caption{MoebileNets clustering on the cat-dog identification task}
    \label{fig:cat_dog_cluster}
\end{figure}

Figure~\ref{fig:MobileNet}c shows to asymmetry measures for human infants, MobileNet with all dogs and cats and Mobile net with the dog breed and cats. We see that the asymmetry is the same for infants and the CNN, indicating a similar effect. As this effect disappears when corrected for by reducing the variance of the dog dataset, we believe that the effect in both infants and CNNs is whole due to image statistics. This finding is backed up by figure~\ref{fig:cat_dog_cluster}, which is the cluster arrangement of the Kaggle `cats vs dogs' dataset in MobileNet. 

\begin{figure}
    \centering
    \includegraphics[width=0.5\textwidth]{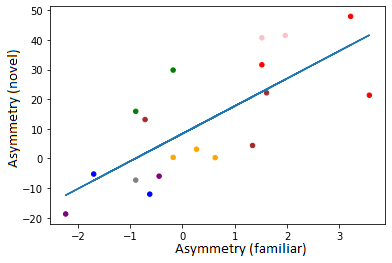}
    \caption{Generally the novel stimulus leads to a a larger asymmetry effect than the familiar stimulus and the more asymmetry of the familar the more the asymmetry of the novel, $R^2=0.55$. Colours refer to types of model, see table~\ref{tab:cat_dog_all_models}.}
    \label{fig:assym_Scores}
\end{figure}

Table~\ref{tab:cat_dog_all_models} and figure~\ref{fig:assym_Scores} show the results for a range of models, most of them agree with MobileNet and have asymmetry towards dogs. We find that asymmetry is always higher on novel stimuli than familiar stimuli across the models, and seems roughly positively correlated with the asymmetry on familiar stimuli and negative correlated with the model size (figure not shown). This indicates that this asymmetry effect is roughly found in CNNs in general, which makes sense if it results from the image statistics as they have all been trained on the same data.  

\begin{table}
    \centering
    \begin{tabular}{|c|r|r|r|c|c|}
    \hline
    NN & Asymmetry on 	& Asymmetry on & No. parameters & Type & Colour\\	
     & familiar stimuli	& novel stimuli & & & \\	
    \hline
    %no_of_params_base_model	type_of_NN	colors
    DenseNet121	& 3.21	& 47.95	& 8,062,504	&1	&red \\%
    DenseNet169	& 1.52	& 31.61	& 14,307,880	&1	&red \\%
    DenseNet201	& 3.57	& 21.34	& 20,242,984	&1	&red\\%
    InceptionResNetV2	& -0.62	& -12.05	&55,873,736	&2	&blue\\%
    InceptionV3	& -1.70	& -5.27& 23,851,784	&2	&blue\\%
    MobileNet	& -0.89	& 15.89&4,253,864	&3	&green\\%
    MobileNetV2	& -0.18	& 29.82&3,538,984	&3	&green\\%
    NASNetLarge	& -2.23	& -18.75&	88,949,818	&4	&purple\\%
    NASNetMobile &	-0.45 & 	-5.98 &	5,326,716	&4	&purple\\%
    ResNet101	& 0.62	& 0.27	& 44,707,176	&5	&orange\\%
    ResNet101V2	& 1.61	& 22.14 &44,675,560	&5	&brown\\%
    ResNet152	& -0.18	& 0.36	&60,419,944	&5	&orange\\%
    ResNet152V2	& 1.34	& 4.37	&60,380,648	&5	&brown\\%
    ResNet50	&0.27	&3.12	&25,636,712	&5	&orange\\%
    ResNet50V2	& -0.71	& 13.12&25,613,800	&5	&brown\\%
    VGG16	& 1.52	& 40.71	&138,357,544	&6	&pink\\%
    VGG19	& 1.96	& 41.52	&143,667,240	&6	&pink\\%
    Xception &	-0.89 &	-7.32&	22,910,480	&2	&gray\\%
    \hline
    \end{tabular}
    \caption{Results from second generation CNNs on the cat-dog identification task. Colour refers to the colour in figure~\ref{fig:assym_Scores}.}
    \label{tab:cat_dog_all_models}
\end{table}

\section{Conclusions}

In this paper we have presented a novel hierarchical clustering algorithm called ClusterFlow, which can cluster high-dimensional data into hypercuboidal clusters based on their position in that space and labels that they possess. We have demonstrated that this clustering works with missing data and that complicated overlapping clusters can be built. The algorithm is fast. We have demonstrated that ClusterFlow can cluster data in lower-dimensional spaces than the one the data lives in, and, by building up clusters into a dimension if and only if a point with a non-zero value in that dimension has been seen, we can create an efficient and conservative exploration of hyperspace. This avoids grabbing lots of dimensions and subspaces where data has not been seen, and it has been demonstrated that this approach removes over-confidence on fooling images, making the system much more resilient to hacking. Due to this conservative subspace expansion, CF is a good algorithm for finding and mapping the manifold that neural networks organise their latent variables onto, and has uses in investigating how NNs work and what they have learned. We then demonstrated that the structure of this manifold is similar to that of a 6 month old infant's for visual recognition of novel and familiar pet based stimuli, showing that modern generation deep-CNNs organise visual information similarly to humans and that CF can crudely map that. Finally, we have shown that using CF as a layer on a pre-trained network we can give the combined algorithm the ability to do relational reasoning on both familiar and completely novel stimuli: this is a huge step forward for A.I. research.

\section{Appendix}

\subsection{Mathematical details}

A \textit{space} is a set of mathematical objects that can be treated as points (although they may not be points) and selected relationships between them (rules for that space). For example, pixels can be thought of as points on a 2-D grid.

%\textit{Manifold}

\textit{A metric space} is a space of points (for example, $\{x,y,z,a,b\}$) defined by a function with the `rules' of a metric, $d$, \emph{distance function} which follows standard distance metric rules:

\begin{equation}
    d(x,y) = 0 \Leftrightarrow x = y \;,
    \label{eq:distance}
\end{equation}
i.e. if the distance between 2 points $x$ and $y$ is zero the points must be the same and \textit{vice versa},
\begin{equation}
    d(x,y) = d(y,x) \;,
\end{equation}
and the triangle inequality:
\begin{equation}
    d(x,z) \leq d(x,y) + d(y,z) \;.
\end{equation}

\subsubsection{Metrics}

\subsubsection{\texorpdfstring{$L^p$}{Lp} norm}

The $L^p$ norm is given by equation~\ref{eq:norm}.

\begin{equation}
    ||\vec{x}||_p = \left( \sum_{i} ||x_i||^p \right)^{1/p}
    \label{eq:norm}
\end{equation}

For example for the vector $\vec{x} = [0, 0, 3, 4, 0]$, $L^0$ is the number of non-zero dimensions and $L^0=2$, $L^1$ is taxi-block distance or the sum so $L^1 = 7$, $L^2$ is the Euclidean distance, so $L^2 = 5$ and the $L^{\infty}$ is defined as the $\mathrm{max}(\vec{x})$, and is 4 in this example.    

%{\displaystyle \|{\boldsymbol {x}}\|_{2}:={\sqrt {x_{1}^{2}+\cdots +x_{n}^{2}}}.}

The Euclidean distance function in a 3-D space satisfies this and is a natural space to describe the 3-D molecular structure in. The Euclidean distance function is also known as the $l^2$ norm, as given by $x = \sqrt{(i^2 + j^2 + k^2)}$ in 3-D space where $x$ is a `point' in that space and $i$, $j$ and $k$ are the \emph{distances} along the Cartesian axes (not the unit vectors, sorry for abuse of terminology). Using the dot product and Euclidean norm you can create a vector space (literally a space where the `points' are vectors), $\mathbb{R}^3$, which is a vector space of 3-tuples of real numbers, the most familiar of which is Cartesian coordinates, $(i, j, k)$ and the vectors are defined from the origin $(0, 0, 0)$.

%\subsection{Mehalanobis distance}

\bibliographystyle{unsrt}  

%%%REFERENCES%%%
\bibliography{references, bristol} %You need to replace "rsc" on this line with the name of your .bib file

\begin{thebibliography}{10}

\bibitem{bronstein2021geometric}
Michael~M Bronstein, Joan Bruna, Taco Cohen, and Petar Veli{\v{c}}kovi{\'c}.
\newblock Geometric deep learning: Grids, groups, graphs, geodesics, and
  gauges.
\newblock {\em arXiv preprint arXiv:2104.13478}, 2021.

\bibitem{40}
Anh Nguyen, Jason Yosinski, and Jeff Clune.
\newblock Deep neural networks are easily fooled: High confidence predictions
  for unrecognizable images.
\newblock In {\em Proceedings of the IEEE Conference on Computer Vision and
  Pattern Recognition}, pages 427--436, 2015.

\bibitem{goodfellow2014explaining}
Ian~J Goodfellow, Jonathon Shlens, and Christian Szegedy.
\newblock Explaining and harnessing adversarial examples.
\newblock {\em arXiv preprint arXiv:1412.6572}, 2014.

\bibitem{61}
Tom~B Brown, Dandelion Man{\'e}, Aurko Roy, Mart{\'\i}n Abadi, and Justin
  Gilmer.
\newblock Adversarial patch.
\newblock {\em arXiv preprint arXiv:1712.09665}, 2017.

\bibitem{41-baker2018deep}
Nicholas Baker, Hongjing Lu, Gennady Erlikhman, and Philip~J Kellman.
\newblock Deep convolutional networks do not classify based on global object
  shape.
\newblock {\em PLoS computational biology}, 14(12):e1006613, 2018.

\bibitem{53}
John~S Denker and Yann {leCun}.
\newblock Transforming neural-net output levels to probability distributions.
\newblock In {\em Advances in neural information processing systems}, pages
  853--859, 1991.

\bibitem{GaleObjectDetectors}
Ella~M Gale, Nicholas Martin, Ryan Blything, Anh Nguyen, and Jeffrey~S Bowers.
\newblock Are there any ‘object detectors’ in the hidden layers of cnns
  trained to identify objects or scenes?
\newblock {\em Vision Research}, 176:60--71, 2020.

\bibitem{57}
Anders Oland, Aayush Bansal, Roger~B Dannenberg, and Bhiksha Raj.
\newblock Be careful what you backpropagate: A case for linear output
  activations \& gradient boosting.
\newblock {\em arXiv preprint arXiv:1707.04199}, 2017.

\bibitem{56}
Chuan Guo, Geoff Pleiss, Yu~Sun, and Kilian~Q Weinberger.
\newblock On calibration of modern neural networks.
\newblock {\em arXiv preprint arXiv:1706.04599}, 2017.

\bibitem{quinn1993evidence}
Paul~C Quinn, Peter~D Eimas, and Stacey~L Rosenkrantz.
\newblock Evidence for representations of perceptually similar natural
  categories by 3-month-old and 4-month-old infants.
\newblock {\em Perception}, 22(4):463--475, 1993.

\bibitem{mareschal2000connectionist}
Denis Mareschal, Robert~M French, and Paul~C Quinn.
\newblock A connectionist account of asymmetric category learning in early
  infancy.
\newblock {\em Developmental psychology}, 36(5):635, 2000.

\bibitem{mareschal2002asymmetric}
Denis Mareschal, Paul~C Quinn, and Robert~M French.
\newblock Asymmetric interference in 3-to 4-month-olds' sequential category
  learning.
\newblock {\em Cognitive Science}, 26(3):377--389, 2002.

\bibitem{32}
David Arthur and Sergei Vassilvitskii.
\newblock k-means++: The advantages of careful seeding.
\newblock In {\em Proceedings of the eighteenth annual ACM-SIAM symposium on
  Discrete algorithms}, pages 1027--1035. Society for Industrial and Applied
  Mathematics, 2007.

\bibitem{murray2016application}
Paul~M Murray, Fiona Bellany, Laure Benhamou, Dejan-Kre{\v{s}}imir Bu{\v{c}}ar,
  Alethea~B Tabor, and Tom~D Sheppard.
\newblock The application of design of experiments (doe) reaction optimisation
  and solvent selection in the development of new synthetic chemistry.
\newblock {\em Organic \& biomolecular chemistry}, 14(8):2373--2384, 2016.

\end{thebibliography}

\end{document}